%% file: main-arxiv.tex
\documentclass{article}

\usepackage{docmute}
\input{settings}

\title{FPGA-Accelerated Correspondence-free Point Cloud Registration with PointNet Features}

\renewcommand{\shorttitle}{FPGA-Accelerated Correspondence-free Point Cloud Registration with PointNet Features}

\author{\href{https://orcid.org/0000-0001-8534-2381}%
  {\includegraphics[scale=0.06]{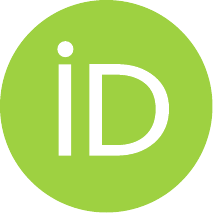}}%
  \hspace{1mm}Keisuke Sugiura\\
  Keio University\\
  3-14-1 Hiyoshi, Kohoku-ku, Yokohama, Japan\\
  \texttt{sugiura@arc.ics.keio.ac.jp}\\
  \And
  \href{https://orcid.org/0000-0001-9578-3842}%
  {\includegraphics[scale=0.06]{orcid.pdf}}%
  \hspace{1mm}Hiroki Matsutani\\
  Keio University\\
  3-14-1 Hiyoshi, Kohoku-ku, Yokohama, Japan\\
  \texttt{matutani@arc.ics.keio.ac.jp}
}

\hypersetup{
  colorlinks = true,
  urlcolor = blue,
  linkcolor = red,
  citecolor = green,
  pdftitle = {FPGA-Accelerated Correspondence-free Point Cloud Registration with PointNet Features},
  pdfsubject = {cs.AR, cs.RO},
  pdfauthor = {Keisuke Sugiura, Hiroki Matsutani},
  pdfkeywords = {Point Cloud Registration, Deep Learning, PointNet, FPGA}
}

\begin{document}

\maketitle

\input{abst}


\input{intro}

\input{related}

\input{preliminaries}

\input{design}

\input{dse}

\input{eval}

\input{conc}


\renewcommand{\baselinestretch}{1.0}
\bibliographystyle{unsrt}

\input{refer}
\vfill

\end{document}

%% file: settings.tex
\usepackage{arxiv}

\usepackage{graphicx}

\usepackage[utf8]{inputenc}
\usepackage[T1]{fontenc}

\usepackage{amsmath}
\usepackage{amsfonts}
\usepackage{bm}
\usepackage{comment}
\usepackage{fancybox}
\usepackage{framed}
\usepackage{color}
\usepackage{multicol}
\usepackage{multirow}
\usepackage{hyperref}
\usepackage{url}

\usepackage{enumitem}

\usepackage{physics}

\usepackage[whole,substmingoth]{bxcjkjatype}

\usepackage{algorithm}
\usepackage[noend]{algpseudocode}
\usepackage{algorithmicx}

\usepackage{tabularx}
\newcolumntype{Y}{>{\centering\arraybackslash}X}

\usepackage{subcaption}

\expandafter\def\expandafter\UrlBreaks\expandafter{\UrlBreaks
  \do\a\do\b\do\c\do\d\do\e\do\f\do\g\do\h\do\i\do\j%
  \do\k\do\l\do\m\do\n\do\o\do\p\do\q\do\r\do\s\do\t%
  \do\u\do\v\do\w\do\x\do\y\do\z\do\A\do\B\do\C\do\D%
  \do\E\do\F\do\G\do\H\do\I\do\J\do\K\do\L\do\M\do\N%
  \do\O\do\P\do\Q\do\R\do\S\do\T\do\U\do\V\do\W\do\X%
  \do\Y\do\Z}

\DeclareMathOperator*{\argmin}{arg\,min}

\DeclareMathOperator{\SE}{\mathrm{SE}}
\DeclareMathOperator{\SO}{\mathrm{SO}}
\DeclareMathOperator{\se}{\mathfrak{se}}

\algrenewcommand\algorithmicindent{1.0em}

\algnewcommand\algorithmicforeach{\textbf{for each}}
\algdef{S}[FOR]{ForEach}[1]{\algorithmicforeach\ #1\ \algorithmicdo}

\algnewcommand\AlgAnd{\textbf{and} }
\algnewcommand\AlgOr{\textbf{or} }
\algnewcommand\AlgContinue{\textbf{Continue}}
\algnewcommand\AlgBreak{\textbf{break}}

\algrenewcommand\textproc{}

\algnewcommand{\Initialize}[1]{
	\State \textbf{Initialize:}
 	\State \hspace*{\algorithmicindent}\parbox[t]{0.8\linewidth}{\raggedright #1}}

\algnewcommand{\LeftComment}[1]{
  \Statex $\triangleright$ #1 \hfill}

\algnewcommand{\IIf}[1]{\State\algorithmicif\ #1\ \algorithmicthen}
\algnewcommand{\EndIIf}{\unskip}

\def\BibTeX{{\rm B\kern-.05em{\sc i\kern-.025em b}\kern-.08em
  T\kern-.1667em\lower.7ex\hbox{E}\kern-.125emX}}


%% file: abst.tex

\begin{abstract}
Point cloud registration serves as a basis for vision and robotic applications including 3D reconstruction and mapping.
Despite significant improvements on the quality of results, recent deep learning approaches are computationally expensive and power-hungry, making them difficult to deploy on resource-constrained edge devices.
To tackle this problem, in this paper, we propose a fast, accurate, and robust registration for low-cost embedded FPGAs.
Based on a parallel and pipelined PointNet feature extractor, we develop custom accelerator cores namely PointLKCore and ReAgentCore, for two different learning-based methods.
They are both correspondence-free and computationally efficient as they avoid the costly feature matching step involving nearest-neighbor search.
The proposed cores are implemented on the Xilinx ZCU104 board and evaluated using both synthetic and real-world datasets, showing the substantial improvements in the trade-offs between runtime and registration quality.
They run 44.08--45.75x faster than ARM Cortex-A53 CPU and offer 1.98--11.13x speedups over Intel Xeon CPU and Nvidia Jetson boards, while consuming less than 1W and achieving 163.11--213.58x energy-efficiency compared to Nvidia GeForce GPU.
The proposed cores are more robust to noise and large initial misalignments than the classical methods and quickly find reasonable solutions in less than 15ms, demonstrating the real-time performance.
\end{abstract}

%% file: intro.tex

\section{Introduction} \label{sec:intro}
Point cloud registration is the key to 3D scene understanding.
It plays a critical role in a wide range of vision and robotic tasks, such as 3D reconstruction~\cite{ShahramIzadi11,RichardANewcombe11}, SLAM~\cite{JiZhang14,TixiaoShan18}, and object pose estimation~\cite{JayMWong17,ChenWang19}.
The registration aims to find a rigid transform (rotation and translation) between two point clouds.
In SLAM, the robot estimates its relative motion by aligning two consecutive LiDAR scans, and also tries to correct the long-term drift by aligning current scans with previous maps when revisiting the same locations (i.e., loop-closure).
The performance of SLAM greatly depends on the underlying registration method.
Ideally, it should be sufficiently accurate and robust, in order to handle real-world scans that are usually perturbed by sensor noise and outliers (e.g., occlusions), and build a consistent map in a large environment.
The energy-efficiency and speed are important factors as well.
Such vision and robotic tasks are usually deployed on mobile edge devices with limited resources and power, and the registration needs to run faster than the data acquisition rate (i.e., process the current scan before the next data arrives).
It is challenging to meet these performance requirements when executed only on embedded CPUs~\cite{LuigiNardi15,KonstantinosBoikos16}, necessitating a fast and energy-efficient registration pipeline with hardware acceleration.

\begin{figure}[htbp]
  \centering
  \includegraphics[keepaspectratio, width=\linewidth]{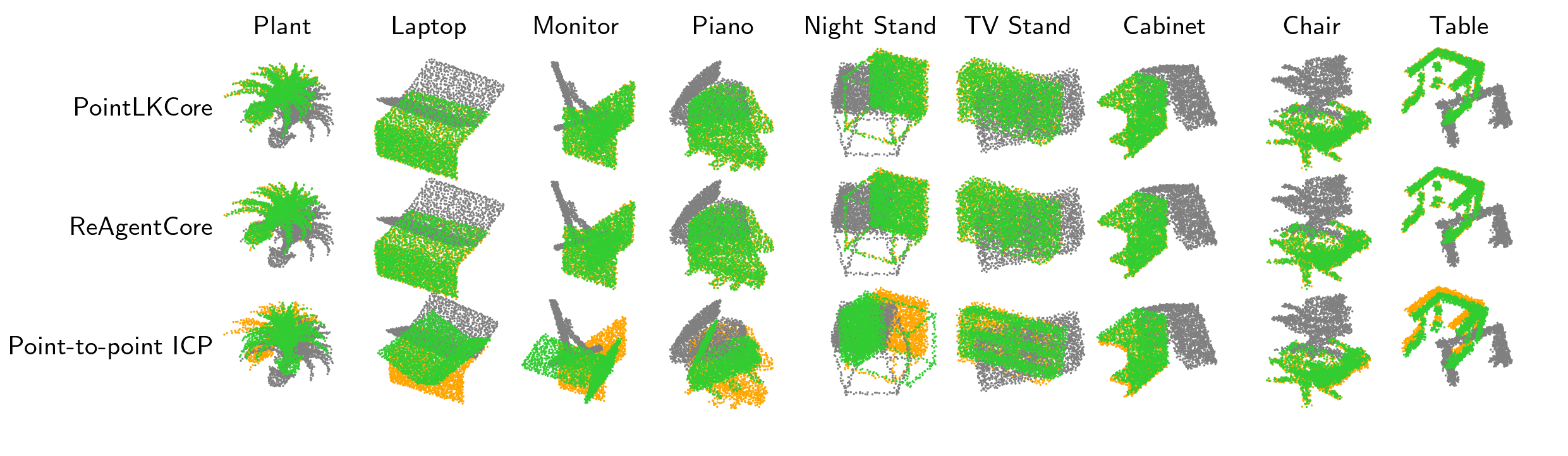}
  \caption{Registration results for ModelNet40 (Unseen) and ScanObjectNN (rightmost three columns) (gray: source, green: transformed source, orange: template).}
  \label{fig:ex7-m40-half2-and-scnn}
\end{figure}

\begin{figure}[htbp]
  \centering
  \includegraphics[keepaspectratio, width=0.9\linewidth]{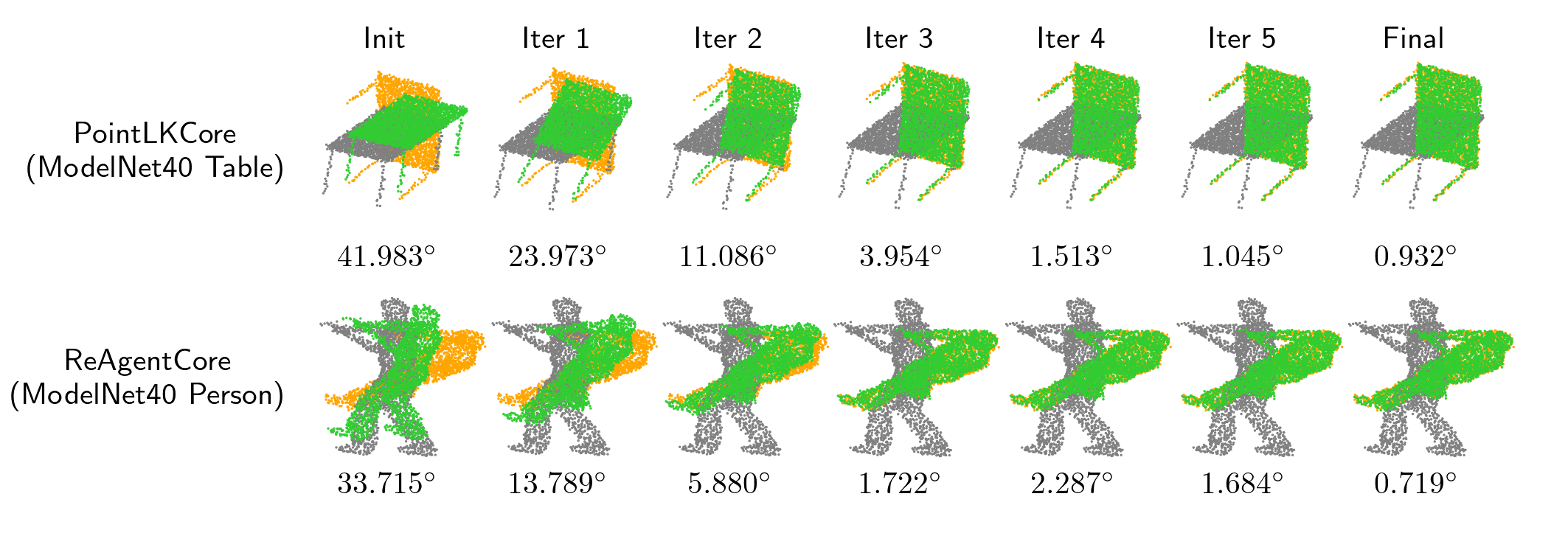}
  \caption{Step-by-step visualization of the registration results (with the rotational ISO (isotropic) errors) (gray: source, green: transformed source, orange: template).}
  \label{fig:ex7-m40-half2-itr}
\end{figure}

Registration is a longstanding research topic.
The widely-known methods, including ICP (Iterative Closest Point)~\cite{Besl92}, RPM (Robust Point Matching)~\cite{StevenGold98}, FGR (Fast Global Registration)~\cite{QianYiZhou16}, and ICP variants~\cite{Segal09,JiaolongYang16,JuyongZhang22} rely on the correspondences between point clouds.
ICP~\cite{Besl92} alternates between establishing point correspondences and computing an alignment that minimizes the distances between matched points using various optimization tools (e.g., SVD (Singular Value Decomposition)~\cite{Besl92} and LM (Levenberg-Marquardt)~\cite{AndrewWFitzgibbon03}).
The former step involves a nearest-neighbor search for every point and has a computational complexity of around $O(N\log N)$ ($N$ is a number of points).
While ICP is fast enough on modern processors, it is susceptible to local minima and cannot provide reliable results without a good initial estimate.
FGR~\cite{QianYiZhou16} uses handcrafted features that encode the local geometry around a point.
While handcrafted features~\cite{AndrewEJohnson99,RaduBogdanRusu08,RaduBogdanRusu09,SamueleSalti14} improve the accuracy of correspondence estimation compared to using the closest-point search, they are computed by simple geometric features (e.g., normals) and are still prone to noise.

Aside from these classical approaches, deep learning-based methods are becoming more prominent in the recent literature, with the aim to extract more distinctive features using dedicated deep neural networks (DNNs)~\cite{AndyZeng17,GilElbaz17,HaowenDeng18A,ChristopherChoy19}.
A common approach is to estimate the soft correspondences (matching probabilities of all possible point pairs) based on the feature similarities, and employ differentiable SVD to compute a rigid transform in one-shot, which makes the registration process fully differentiable and thus end-to-end trainable~\cite{YueWang19A,YueWang19B,ZiJianYew20,ChristopherChoy20,MohamedElBanani21}.
The soft correspondence leads to $O(N^2)$ complexity and may not be applicable to large-scale or dense point clouds.
While the learned feature representation has yielded significant improvements in the accuracy and robustness, learning-based methods are more computationally and memory demanding than the classical approaches due to a large number of parameters and operations.

Several learning-based methods avoid the costly correspondence search using different formulations.
PointNetLK~\cite{YasuhiroAoki19} and ReAgent~\cite{DominikBauer21} utilize the global features extracted by PointNet~\cite{CharlesRQi17} that describe the entire point clouds.
PointNetLK~\cite{YasuhiroAoki19} is a seminal work that applies the iterative LK (Lucas-Kanade) optimization~\cite{SimonBaker04}; it aligns two point clouds by minimizing the residual between two PointNet features.
ReAgent~\cite{DominikBauer21} treats the registration as a multi-class classification problem; at each step, it predicts a discrete action label that maximally reduces the registration error.
These correspondence-free methods are inherently efficient, because PointNet inference only requires $O(N)$ computational and memory cost, making them suitable candidates for real-time application on edge devices.

In this paper, we propose an efficient point cloud registration pipeline for embedded FPGAs.
We first design a PointNet feature extractor module with a pipeline architecture, which consumes only $O(1)$ on-chip memory thanks to the simplified PointNet architecture and modified feature extraction flow.
On top of this, we develop two customizable accelerator cores for PointNetLK and ReAgent, namely \textbf{PointLKCore} and \textbf{ReAgentCore}.
To save the on-chip memory, we apply the recently-proposed LLT (Learnable Lookup Table)~\cite{LongguangWang22} quantization to the network, which requires little additional overhead for lookup operations.
All network parameters can be stored on-chip as a result, eliminating most of the off-chip memory accesses.
While registration is a complicated task involving feature extraction and various geometric operations, we build accurate models for the clock-cycle latency and resource utilization to conduct design-space exploration.
We evaluate the proposed cores on the Xilinx ZCU104 board using both synthetic ModelNet40~\cite{ZhirongWu15} and real-world ScanObjectNN~\cite{MikaelaAngelinaUy19} datasets.
The experimental results confirm a significant improvement in the trade-off between runtime and registration quality.
Our contributions are summarized as follows:
\begin{enumerate}[leftmargin=*]
  \item To the best of our knowledge, we are the first to introduce FPGA accelerators for the deep learning-based point cloud registration.
  The proposed accelerator cores utilize the parallel and pipelined PointNet feature extractor module.
  We simplify the PointNet architecture and modify the feature extraction algorithm, such that the on-chip memory cost remains constant regardless of the input size, leading to resource-efficiency and scalability.
  \item We develop accurate performance models for the proposed accelerators.
  Based on these, we conduct the design-space exploration to fully harness the available resources on a specified FPGA board and minimize the latency.
  \item For resource-efficiency, we apply the low-overhead lookup-table quantization~\cite{LongguangWang22} to the network parameters.
  While it is previously applied to the famous semantic tasks (e.g., classification and segmentation), we show its effectiveness in the geometric tasks for the first time.
  \item For PointNetLK, we present a simple approach for Jacobian computation to further improve the accuracy.
  Instead of the backward difference, we opt to use the central difference approximation to compute the gradient of PointNet features with respect to transform parameters.
  This yields on-par or better accuracy while being significantly faster than using the analytical formulation~\cite{XueqianLi21}.
\end{enumerate}

%% file: related.tex

\section{Related Work} \label{sec:related}
\subsection{Deep Learning-based Point Cloud Registration} \label{sec:related-dnn-registration}
\subsubsection{Correspondence Approach} \label{sec:related-correspondence-approach}
The long-established approach for registration is to use the correspondences, i.e., compute local descriptors for each point, match these descriptors via nearest-neighbor search to establish correspondences, and recover the rigid transform between two point clouds in one-shot (e.g., SVD) or using robust estimators (e.g., RANSAC).
A line of work has focused on extracting more distinctive features using DNNs for reliable feature matching~\cite{AndyZeng17,GilElbaz17,MarcKhoury17,HaowenDeng18A,HaowenDeng18B,ChristopherChoy19,XuyangBai20,ShengyuHuang21,HaoYu21,YangLi22,HaipingWang22}.
3DMatch~\cite{AndyZeng17} uses a 3D CNN model to obtain features from voxel grids representing local surface patches, whereas PPFNet~\cite{HaowenDeng18A} learns global context-aware features using PointNet~\cite{CharlesRQi17}.
Despite the steady development, these correspondence-based methods are subject to outliers, which are often the case if point clouds have symmetric/repetitive structures (e.g., shelf) or no salient structures (e.g., smooth surface).
Besides, robust estimators require a large number of trials to reject incorrect matches, which slows down the registration process~\cite{ZiJianYew22}.

Several works combine feature extraction and rigid transform estimation into an end-to-end trainable framework~\cite{YueWang19A,YueWang19B,ZiJianYew20,ChristopherChoy20,TaewonMin21}.
While ICP assigns each point in one point cloud to the closest point in the other (i.e., hard assignment), DCP~\cite{YueWang19A} predicts soft correspondences (matching probabilities between all point pairs) based on the feature similarities and perform differentiable SVD to obtain rigid transforms.
It employs a graph convolution- and attention-based network that encode both intra- and inter-point cloud information.
PRNet~\cite{YueWang19B} modifies the softmax operation in DCP to adaptively control the sharpness of the matching.
RPM-Net~\cite{ZiJianYew20} extends RPM~\cite{StevenGold98} by using learned feature distances instead of spatial distances and introducing a differentiable Sinkhorn normalization.
Because soft correspondences have $O(N^2)$ complexity, these methods may not scale to large-scale point clouds.
To avoid the costly feature matching, DeepVCP~\cite{WeixinLu19} generates virtual points from neighboring points using learned weights and establishes one-to-one correspondences.
CorsNet~\cite{AkiyoshiKurobe20} tries to match two point clouds by moving points in one point cloud using predicted 3D translations.
RegTR~\cite{ZiJianYew22} follows a similar idea; it uses a stack of self- and cross-attention layers to extract point features, which are then passed to an MLP to generate point coordinates.
While attention mechanism allows to incorporate contextual information into features, it has a quadratic computational complexity and suffers from the lack of scalability.

\subsubsection{Regression or Classification Approach} \label{sec:related-regression-approach}
Aside from the above correspondence-based methods, there exist abundant studies that employ DNNs for transform estimation as well as for feature extraction~\cite{JiaxinLi17,MichelleValente19,LiDing19,VinitSarode19,GDiasPais20,HaoXu21,DonghoonLee21}.
Li \textit{et al.}~\cite{JiaxinLi17} present a dual-purpose CNN model for scan matching and loop closure detection in 2D LiDAR SLAM, while Valente \textit{et al.}~\cite{MichelleValente19} use a CNN- and LSTM-based model to capture the temporal features of 2D LiDAR scans.
DeepMapping~\cite{LiDing19} is a framework to align a sequence of 2D LiDAR scans and build a consistent occupancy grid map via unsupervised training.
PCRNet~\cite{VinitSarode19} uses an MLP to recover transform parameters (a quaternion and translation) from a pair of global features extracted by a Siamese PointNet.
One drawback of such methods is that it is generally challenging to accurately regress the transform parameters, since DNN models need to learn the properties of rotation representations\footnote{For example, a quaternion should be unit-length and keep its scalar component positive to avoid ambiguity.
Euler angles suffer from the discontinuities and singularities.}.
ReAgent~\cite{DominikBauer21} takes a unique approach that benefits from IL (imitation learning) and RL (reinforcement learning) techniques.
It divides the translation and rotation angles into discrete bins (i.e., actions) and poses the registration task as a classification problem, which is easier than the direct regression.

\subsubsection{Lucas-Kanade and Direct Feature Alignment Approach} \label{sec:related-lk-approach}
Another approach is to directly align global features that encode the whole point clouds by an iterative LK (Lucas-Kanade) optimization~\cite{SimonBaker04}.
PointNetLK~\cite{YasuhiroAoki19} extracts global features using PointNet~\cite{CharlesRQi17} and computes a rigid transform that minimizes the feature difference.
The rationale behind this is that PointNet should produce similar features if two point clouds are closely aligned to each other.
PointNetLK relies on a finite difference approximation to compute a Jacobian of the PointNet feature with respect to transform parameters.
Sekikawa \textit{et al.}~\cite{YusukeSekikawa19} replace MLPs with lookup tables to eliminate vector-matrix operations and speed up the PointNet feature extraction.
FMR~\cite{XiaoshuiHuang20} is a simple extension to PointNetLK; it adds a decoder block to the PointNet to extract more distinctive features and allow either semi-supervised or unsupervised training.
Li \textit{et al.}~\cite{XueqianLi21} derive an analytical Jacobian consisting of two terms (feature gradient and warp Jacobian) to avoid numerical instabilities and improve generalizability.
Importantly, PointNetLK and ReAgent are both correspondence-free as they focus on the global representation of point clouds rather than the local geometry around each point.
They circumvent the costly NN (nearest neighbor) search and can be characterized by the lower ($O(N)$) computational cost.
We opt to use them as a backbone for the efficient point cloud registration on embedded FPGAs.

\subsection{FPGA-based Acceleration of Point Cloud Registration} \label{sec:related-fpga-registration}
Despite of the importance and broad application, the FPGA acceleration of point cloud registration has yet to be fully explored.
Kosuge \textit{et al.}~\cite{AtsutakeKosuge20} propose an ICP accelerator for object pose estimation, which is a core functionality in picking robots.
They use the hierarchical graph instead of K-d tree for improved $k$NN ($k$-nearest neighbor) search efficiency, and their accelerator performs the distance computation and sorting in parallel for graph generation and $k$NN.
$k$NN becomes a performance bottleneck in ICP and its acceleration is still under ongoing research.
Belshaw \textit{et al.}~\cite{MichaelSBelshaw09} parallelize the brute-force NN for ICP-based object tracking, Sun \textit{et al.}~\cite{HaoSun20} devise a voxel-based two-layer data structure for the registration of LiDAR scans in 3D SLAM~\cite{JiZhang14}, and Li \textit{et al.}~\cite{YimingLi22} present a $k$NN accelerator based on the approximate K-d tree, which consists of the parallel merge sorting and distance computation units.
Deng \textit{et al.}~\cite{QiDeng21} introduce an FPGA accelerator for NDT (Normal Distributions Transform) by utilizing a non-recursive voxel data structure.
NDT~\cite{Biber03} splits the point cloud into a set of voxels, with each modeled as a normal distribution of points that lie inside it.
The authors of \cite{KeisukeSugiura22} propose an accelerator for the registration between a 2D LiDAR scan and an occupancy grid map, which is applied to various 2D SLAM methods.
In \cite{MarcEisoldt21,MarcelFlottmann21}, the authors focus on the TSDF (Truncated Signed Distance Function)-based 3D SLAM and implement the registration and map update steps on FPGA.
These works successfully demonstrate the effectiveness of FPGA acceleration for the non-learning-based methods, while they are often sensitive to the initial guesses and susceptible to local minima.
Compared to these, we put a focus on the deep learning-based methods; they offer better accuracy and robustness to noise, and are well-suited to FPGAs owing to the massive parallelism of DNNs.

Compared to our previous work~\cite{KeisukeSugiura23}, where we only implement the PointNet feature extraction part on FPGA to accelerate the PointNetLK registration, this paper makes the following improvements.
We exploit more parallelism in the feature extraction (e.g., process multiple points in parallel) and present two newly-designed unified accelerator cores that fully implement PointNetLK and ReAgent.
The network parameters are stored on-chip thanks to the simple network architecture and LLT quantization~\cite{LongguangWang22}.
We build accurate resource models and conduct design-space exploration to find optimal design parameters.
For PointNetLK, we introduce a simple yet effective Jacobian computation method and jointly train the model with a classifier or decoder branch.
In addition to embedded CPUs (ARM Cortex-A53), we compare the proposed cores with embedded GPUs (Nvidia Jetson) and a desktop computer (Intel CPU and Nvidia GeForce GPU) to highlight the performance benefits of our approach.

%% file: preliminaries.tex

\section{Preliminaries} \label{sec:prelim}
\subsection{Problem Formulation} \label{sec:prelim-problem-formulation}
Given a source and template $\mathcal{P}_S, \mathcal{P}_T \in \mathbb{R}^{N \times 3}$ containing $N$ points each\footnote{For simplicity, we assume that source and template have the same number of points.}, the registration seeks to find a rigid transform $\vb{G} = \left[ \vb{R} \mid \vb{t} \right] \in \SE(3)$ that best aligns $\mathcal{P}_S$ with $\mathcal{P}_T$, where $\vb{R} \in \SO(3)$ and $\vb{t} \in \mathbb{R}^3$ denote a rotation and translation.
One typical approach is to establish the correspondences between $\mathcal{P}_S$ and $\mathcal{P}_T$, e.g., by finding a closest point in $\mathcal{P}_T$ for each point in $\mathcal{P}_S$, but it involves a costly NN search.
Besides, some points in $\mathcal{P}_S$ may not have matching points in $\mathcal{P}_T$ due to the different number of points or density distribution; the noise and occlusion break the point correspondences as well.
The presence of symmetric and repetitive structures in point clouds leads to unreliable or incorrect matches (outliers).
On the other hand, both PointNetLK and ReAgent are correspondence-free and therefore avoid these issues; they instead rely on the global features extracted by PointNet.
They only take point coordinates as input and do not require other geometric features such as surface normals, which eliminates the preprocessing cost.
We briefly describe PointNetLK and ReAgent in the following.

\subsection{PointNetLK} \label{sec:prelim-pointlk}
The method is summarized in Alg. \ref{alg:pointlk}.
We denote by $\vb*{\phi}(\mathcal{P}): \mathbb{R}^{N \times 3} \to \mathbb{R}^K$ PointNet that encodes a point cloud into a $K$-dimensional global feature vector.
PointNetLK tries to minimize the error $\mathcal{L}_\mathrm{feat}(\vb{G})$ between two global features, $\vb*{\phi}(\vb{G} \cdot \mathcal{P}_S)$ and $\vb*{\phi}(\mathcal{P}_T)$, instead of spatial distances between matched point pairs as in ICP.
The key idea is that PointNet should produce similar features if two point clouds are well-aligned.
Note that $\vb{G}(\vb*{\xi}) = \exp(\vb*{\xi}^\wedge)$ is recovered from a 6D twist parameter $\vb*{\xi} \in \mathbb{R}^6$ via exponential map, and $\wedge$ is a wedge operator~\cite{Barfoot17} which maps from $\mathbb{R}^6$ to $\se(3)$ Lie algebra.
The registration problem is thus formulated as:
\begin{equation}
  \vb*{\xi}^* = \argmin_{\vb*{\xi}} \mathcal{L}_\mathrm{feat}(\vb{G}(\vb*{\xi}))
    = \argmin_{\vb*{\xi}} \left\| \vb*{\phi}(\mathbf{G}(\vb*{\xi}) \cdot \mathcal{P}_S)
    - \vb*{\phi}(\mathcal{P}_T) \right\|^2.
  \label{eq:pointlk-objective-naive}
\end{equation}
PointNetLK employs the IC (inverse-compositional) formulation and swap the roles of template and source, i.e., it solves for $\vb*{\xi}$ such that its inverse $\vb{G}(\vb*{\xi})^{-1} = \exp(-\vb*{\xi}^\wedge)$ best aligns $\mathcal{P}_T$ with $\mathcal{P}_S$:
\begin{equation}
  \vb*{\xi}^* = \argmin_{\vb*{\xi}} \left\| \vb*{\phi}(\mathcal{P}_S)
    - \vb*{\phi}(\mathbf{G}(\vb*{\xi})^{-1} \cdot \mathcal{P}_T) \right\|^2.
  \label{eq:pointlk-objective}
\end{equation}
$\vb{G}(\vb*{\xi})$ is updated as $\vb{G}_i \gets \vb{\Delta}\vb{G}_i \cdot \vb{G}_{i - 1}$, where $i$ denotes the iteration.
The twist parameter $\vb{\Delta}\vb*{\xi}_i$ for the incremental transform $\vb{\Delta}\vb{G}_i = \exp(\vb{\Delta}\vb*{\xi}_i^\wedge)$ satisfies:
\begin{equation}
  \vb{\Delta}\vb*{\xi}_i^* = \argmin_{\vb{\Delta}\vb*{\xi}_i} \left\|
    \vb*{\phi}(\vb{G}_{i - 1} \cdot \mathcal{P}_S)
    - \vb*{\phi}(\vb{\Delta} \vb{G}_i^{-1} \cdot \mathcal{P}_T) \right\|^2
  \simeq \argmin_{\vb{\Delta}\vb*{\xi}_i} \left\|
    \vb*{\phi}(\vb{G}_{i - 1} \cdot \mathcal{P}_S)
    - \vb*{\phi}(\mathcal{P}_T) - \vb{J} \vb{\Delta}\vb*{\xi}_i \right\|^2.
  \label{eq:pointlk-objective-linearized}
\end{equation}
In Eq. \ref{eq:pointlk-objective-linearized}, the feature residual is linearized at $\vb{\Delta}\vb*{\xi}_i = \vb{0}$ by Taylor expansion.
The Jacobian $\vb{J} \in \mathbb{R}^{K \times 6}$ represents how the PointNet feature $\vb*{\phi}(\mathcal{P}_T)$ changes with respect to the pose, which is defined as ($\vb{\Delta}\vb{G}_i^{-1} = \exp(-\vb{\Delta}\vb*{\xi}_i^\wedge)$):
\begin{equation}
  \vb{J} =  \left. \pdv{\vb{\Delta}\vb*{\xi}^\top}
    \vb*{\phi}(\exp(-\vb{\Delta}\vb*{\xi}^\wedge) \cdot \mathcal{P}_T)
    \right|_{\vb{\Delta}\vb*{\xi} = \vb{0}}
  \label{eq:pointlk-jacobian}
\end{equation}
$\vb{J}$ is approximated by the (backward) finite difference.
Its $j$-th column is written as ($j = 1, \ldots, 6$):
\begin{equation}
  \vb{J}_j = \frac{1}{t_j} \left( \vb*{\phi}(\vb*{\delta}\vb{G}_j^- \cdot \mathcal{P}_T)
    - \vb*{\phi}(\mathcal{P}_T) \right)
    \quad (\vb*{\delta}\vb{G}_j^\pm = \exp(\pm t_j \vb{e}_j^\wedge)),
  \label{eq:pointlk-jacobian-backward}
\end{equation}
where $t_j$ denotes an infinitesimal step (e.g., $10^{-2}$) and $\vb{e}_j \in \mathbb{R}^6$ is a unit vector with one for the $j$-th element and zeros elsewhere.
The Jacobian computation is expensive, as PointNetLK needs to perturb the template and extract a perturbed feature $\vb*{\phi}(\vb*{\delta}\vb{G}_j^- \cdot \mathcal{P}_T)$ six times in total.
Taking a partial derivative of Eq. \ref{eq:pointlk-objective-linearized} with respect to $\vb{\Delta}\vb*{\xi}_i$ and setting it to zero yields the optimal twist $\vb{\Delta}\vb*{\xi}_i^*$:
\begin{equation}
  \vb{\Delta}\vb*{\xi}_i^* = \vb{J}^\dagger \left(
    \vb*{\phi}(\vb{G}_{i - 1} \cdot \mathcal{P}_S) - \vb*{\phi}(\mathcal{P}_T) \right),
  \label{eq:pointlk-solution}
\end{equation}
where $\vb{J}^\dagger = (\vb{J}^\top \vb{J})^{-1} \vb{J}^\top \in \mathbb{R}^{6 \times K}$ is a pseudoinverse of $\vb{J}$.
The algorithm is outlined as follows: at initialization, PointNetLK computes a pseudoinverse of the Jacobian (Alg. \ref{alg:pointlk}, lines \ref{alg:pointlk-init-begin}--\ref{alg:pointlk-init-end}).
Then, it proceeds to the iterative LK optimization (lines \ref{alg:pointlk-opt-begin}--\ref{alg:pointlk-opt-end}).
It transforms the source by the current estimate $\vb{G}_{i - 1}$ and extracts a feature $\vb*{\phi}(\vb{G}_{i - 1} \cdot \mathcal{P}_S)$ (line \ref{alg:pointlk-encode-source}).
Using $\vb{J}^\dagger$ and Eq. \ref{eq:pointlk-solution}, it computes an update $\vb{\Delta}\vb*{\xi}_i^*$ to obtain a new estimate ($\vb{G}_i \gets \exp(\vb{\Delta}\vb*{\xi}_i^{*\wedge}) \cdot \vb{G}_{i - 1}$) (lines \ref{alg:pointlk-compute-twist}--\ref{alg:pointlk-transform-update}).
This process is repeated until convergence ($\left\| \vb{\Delta}\vb*{\xi}_i^* \right\| < \varepsilon$) or the maximum number of iterations $I_{\max}$ is reached.
Note that $\vb{J}$ in Eq. \ref{eq:pointlk-jacobian-backward} does not depend on the index $i$, meaning that $\vb{J}$ and $\vb{J}^\dagger$ are precomputed only once and fixed throughout the iterations.
IC formulation hence greatly reduces the computational cost; in the original formulation, $\vb{J}$ is a gradient of the source feature $\vb*{\phi}(\vb{G}_i \cdot \mathcal{P}_S)$ and hence needs to be recomputed at every iteration.

\begin{algorithm}[h]
  \caption{Point cloud registration with PointNetLK}
  \label{alg:pointlk}
  \begin{algorithmic}[1]
    \Require Source $\mathcal{P}_S$, template $\mathcal{P}_T$,
      initial transform $\vb{G}_0 = \vb{I}$ (identity), PointNet $\vb*{\phi}$
    \Ensure Rigid transform $\vb{G} \in \SE(3)$ from $\mathcal{P}_S$ to $\mathcal{P}_T$
      \vspace*{2.5pt}
    \LeftComment{\textbf{Initialization (Jacobian computation)}}
    \State Compute a global feature of template: $\vb*{\phi}(\mathcal{P}_T) \in \mathbb{R}^K$
      \label{alg:pointlk-encode-template} \label{alg:pointlk-init-begin}
    \State Perturb a template six times: $\{ \vb*{\phi}(\vb*{\delta}\vb{G}_j^- \cdot \mathcal{P}_T) \}, \ j = 1, \ldots, 6$
      \label{alg:pointlk-perturb-template}
    \State Compute a Jacobian: $\vb{J} \in \mathbb{R}^{K \times 6}$
      (Eq. \ref{eq:pointlk-jacobian-backward})
      \label{alg:pointlk-compute-jacobian}
    \State Compute a pseudoinverse of Jacobian: $\vb{J}^\dagger = (\vb{J}^\top \vb{J})^{-1} \vb{J}^\top \in \mathbb{R}^{6 \times K}$
      \label{alg:pointlk-compute-inverse} \vspace*{2.5pt}
      \label{alg:pointlk-init-end}
    \LeftComment{\textbf{Iterative optimization (Lucas-Kanade)}}
    \For{$i = 1, 2, \ldots, I_{\max}$} \label{alg:pointlk-opt-begin}
      \State Compute a global feature of source: $\vb*{\phi}(\vb{G}_{i - 1} \cdot \mathcal{P}_S)$
        \label{alg:pointlk-encode-source}
      \State Compute an optimal twist:
        $\vb{\Delta}\vb*{\xi}_i \gets \vb{J}^\dagger \left(
        \vb*{\phi}(\vb{G}_{i - 1} \cdot \mathcal{P}_S) - \vb*{\phi}(\mathcal{P}_T) \right)$
        \label{alg:pointlk-compute-twist}
      \State Update the rigid transform: $\vb{G}_i \gets
        \exp(\vb{\Delta}\vb*{\xi}_i^\wedge) \cdot \vb{G}_{i - 1}$
        \label{alg:pointlk-transform-update}
      \IIf{$\left\| \vb{\Delta}\vb*{\xi}_i \right\| < \varepsilon$}
        \AlgBreak \Comment{Check convergence}
        \label{alg:pointlk-check-convergence}
      \EndIIf
    \EndFor
    \State \Return $\vb{G}_i$ \label{alg:pointlk-opt-end}
  \end{algorithmic}
\end{algorithm}

\subsection{ReAgent} \label{sec:prelim-reagent}
Similar to PointNetLK, ReAgent is an iterative method and uses PointNet for feature extraction.
Alg. \ref{alg:reagent} presents the algorithm.
At iteration $i$, it computes a source feature, $\vb*{\phi}(\vb{G}_{i - 1} \cdot \mathcal{P}_S)$, which is concatenated with a precomputed template feature $\vb*{\phi}(\mathcal{P}_T)$ to form a $2K$-dimensional state vector $\vb*{s}_i = (\vb*{\phi}(\vb{G}_{i - 1} \cdot \mathcal{P}_S), \vb*{\phi}(\mathcal{P}_T))$ (Alg. \ref{alg:reagent}, line \ref{alg:reagent-encode-source}).
ReAgent employs two actor networks to determine the translational and rotational actions (i.e., step sizes, line \ref{alg:reagent-action}).
Specifically, each actor takes $\vb*{s}_i$ as input and produces an output of size $(3, 2N_\mathrm{act} + 1)$, containing probabilities of $2N_\mathrm{act} + 1$ possible actions for each translational or rotational axis (i.e., degree of freedom).
$N_\mathrm{act}$ is a hyperparameter and set to 5.
The actions with the largest probabilities yield two action vectors, $\vb{a}_i^t = (a_{i, x}^t, a_{i, y}^t, a_{i, z}^t)$ and $\vb{a}_i^r = (a_{i, x}^r, a_{i, y}^r, a_{i, z}^r)$, with each element in the range of $[0, 2N_\mathrm{act}]$.
The update $\vb{\Delta}\vb{G}_i = [\vb{R}(\vb{a}_i^r) \mid \vb{t}(\vb{a}_i^t)] \in \SE(3)$ is obtained as:
\begin{equation}
  \vb{t}(\vb{a}_i^t) = [\mathbb{T}(a_{i, x}^t),
    \mathbb{T}(a_{i, y}^t), \mathbb{T}(a_{i, z}^t)]^\top, \
  \vb{R}(\vb{a}_i^r) = \vb{R}_x(\mathbb{T}(a_{i, x}^r))
    \vb{R}_y(\mathbb{T}(a_{i, y}^r)) \vb{R}_z(\mathbb{T}(a_{i, z}^r)),
  \label{eq:reagent-action-to-transform}
\end{equation}
where $\vb{R}_{\{x, y, z\}}(\theta)$ represents a rotation around the respective axis by an angle $\theta$, and the table $\mathbb{T}$ maps action labels to the corresponding step sizes.
ReAgent uses exponential step sizes defined as:
\begin{equation}
  \mathbb{T}(a) = 0 \ (a = N_\mathrm{act}), \quad
  -(1 / 900) \cdot 3^{N_\mathrm{act} - a} \ (0 \le a < N_\mathrm{act}), \quad
  (1 / 900) \cdot 3^{a - N_\mathrm{act}} \ (N_\mathrm{act} < a \le 2N_\mathrm{act}).
  \label{eq:reagent-step-sizes}
\end{equation}
Using $\vb{\Delta}\vb{G}_i$, ReAgent updates $\vb{G}$ in a disentangled manner (line \ref{alg:reagent-transform-update}).
Instead of the standard composition, i.e., $\vb{G}_i \gets \vb{\Delta}\vb{G}_i \cdot \vb{G}_{i - 1}$ ($\vb{R}_i = \vb{R}(\vb{a}_i^r) \vb{R}_{i - 1}$ and $\vb{t}_i = \vb{t}(\vb{a}_i^t) + \vb{R}_i(\vb{a}_i^r) \vb{t}_{i - 1}$), the new transform $\vb{G}_i = [\vb{R}_i \mid \vb{t}_i]$ is computed as\footnote{
Using such disentangled form $\vb{G} = [\vb{R} \mid \vb{t}]$, the point cloud $\mathcal{P}$ is transformed as $\vb{R}(\mathcal{P} - \vb*{\mu}) + \vb*{\mu} + \vb{t}$, i.e., $\mathcal{P}$ is first zero-centered by translating its centroid $\vb*{\mu}$ to the origin and rotated by $\vb{R}$.
It is moved back to the original position and translated by $\vb{t}$.}:
\begin{equation}
  \vb{R}_i = \vb{R}(\vb{a}_i^r) \vb{R}_{i - 1}, \ \vb{t}_i = \vb{t}(\vb{a}_i^t) + \vb{t}_{i - 1}.
  \label{eq:reagent-transform-update}
\end{equation}
In this way, $\vb{t}_i$ is updated without the rotation $\vb{R}_i(\vb{a}_i^r)$; the actor network therefore needs to account for either pure translation or rotation, which leads to the improved accuracy and interpretability.
ReAgent continues to the next iteration until the maximum number of iterations $I_{\max}$.

While PointNetLK treats the update $\vb{\Delta}\vb{G}$ as a continuous variable, ReAgent computes $\vb{\Delta}\vb{G}$ based on the discrete step sizes and casts the registration task as an iterative classification problem.
ReAgent is trained end-to-end using IL, i.e., actor networks learn to produce optimal action labels that maximally reduce the registration error by imitating the expert demonstration.
RL can also be used by designing a reward function that penalizes actions leading to a higher error.

\begin{algorithm}[h]
  \caption{Point cloud registration with ReAgent}
  \label{alg:reagent}
  \begin{algorithmic}[1]
    \Require Source $\mathcal{P}_S$, template $\mathcal{P}_T$,
      initial transform $\vb{G}_0 = \vb{I}$ (identity), PointNet $\vb*{\phi}$
    \Ensure Rigid transform $\vb{G} \in \SE(3)$ from $\mathcal{P}_S$ to $\mathcal{P}_T$
      \vspace*{2.5pt}
    \State Compute a global feature of template: $\vb*{\phi}(\mathcal{P}_T) \in \mathbb{R}^K$
      \label{alg:reagent-encode-template}
    \For{$i = 1, 2, \ldots, I_{\max}$}
      \State Compute a global feature of source: $\vb*{\phi}(\vb{G}_{i - 1} \cdot \mathcal{P}_S)$
        \label{alg:reagent-encode-source}
      \State Determine translational and rotational actions using actor networks: $\vb{a}_i^t, \vb{a}_i^r$
        \label{alg:reagent-action}
      \State Compute an update $\vb{\Delta}\vb{G}_i = [\vb{R}(\vb{a}_i^r) \mid \vb{t}(\vb{a}_i^t)]$ (Eq. \ref{eq:reagent-action-to-transform}) and a new transform $\vb{G}_i$ (Eq. \ref{eq:reagent-transform-update})
        \label{alg:reagent-transform-update}
    \EndFor
    \State \Return $\vb{G}_i$
  \end{algorithmic}
\end{algorithm}

%% file: design.tex

\section{Design of Registration Accelerators} \label{sec:design}
In this section, we propose a lightweight PointNet feature extractor, based on which we design accelerator IP cores, namely \textbf{PointLKCore} and \textbf{ReAgentCore}, for two deep learning-based registration methods.

\subsection{Design of the Point Cloud Feature Extractor} \label{sec:design-feature-extractor}
Feature extraction is a key step in the learning-based registration and forms a large portion of the computation time (Fig. \ref{fig:ex3-break-m40-table}).
To cope with its computational complexity, we first design a pipelined and parallelized feature extractor module.

\begin{figure}[h]
  \centering
  \includegraphics[keepaspectratio, width=0.6\linewidth]{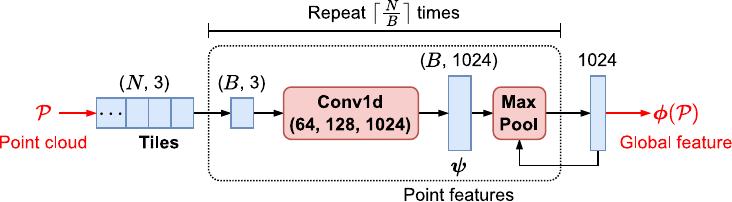}
  \caption{Overview of the PointNet feature extractor module.
  $N$ points are processed in tiles of $B$ points to reduce the on-chip memory cost (for intermediate point features) from $O(N)$ to $O(B)$.}
  \label{fig:network-model-pointnet1}
\end{figure}

The module extracts a global feature $\vb*{\phi}(\mathcal{P})$ using PointNet for a given point cloud $\mathcal{P}$.
As shown in Fig. \ref{fig:network-model-pointnet1}, the network is divided into two parts: pointwise feature extraction and aggregation.
It first computes 1024D point features $\vb{\Psi} = \left\{ \vb*{\psi}(\vb{p}_1), \ldots, \vb*{\psi}(\vb{p}_N) \right\} \in \mathbb{R}^{N \times 1024}$ for $N$ input points $\mathcal{P} = \left\{ \vb{p}_1, \ldots, \vb{p}_N \right\} \in \mathbb{R}^{N \times 3}$ using three 1D convolution layers with output dimensions of $(64, 128, 1024)$\footnote{Since the kernel size and stride are fixed to one, the 1D convolution is equivalent to a matrix product and fully-connected layers.}.
These point features are then aggregated into a global feature $\vb*{\phi}(\mathcal{P}) = \max(\vb*{\psi}(\vb{p}_1), \ldots, \vb*{\psi}(\vb{p}_N))$ by the last max-pooling layer.
While $\vb*{\phi}(\mathcal{P})$ is obtained with one forward pass of PointNet as above, this standard approach requires $O(N)$ memory space to store the intermediate point features of size $N \times n$ ($n = 64, 128, 1024$).
Instead of the above, the module processes $N$ points in tiles of size $B$ and computes $\vb*{\phi}(\mathcal{P})$ as follows (Fig. \ref{fig:network-model-pointnet1}).
At initialization, $\vb*{\phi}(\mathcal{P})$ is set to $-\infty$.
It then (1) retrieves a new tile $\left\{ \vb{p}_i, \ldots, \vb{p}_{i + B - 1} \right\}$ from the external memory and (2) transforms it into 1024D point features $\left\{ \vb*{\psi}(\vb{p}_i), \ldots, \vb*{\psi}(\vb{p}_{i + B - 1}) \right\}$ using convolution layers.
It (3) updates the global feature via max-pooling: $\vb*{\phi}(\mathcal{P}) \gets \max(\vb*{\phi}(\mathcal{P}), \vb*{\psi}(\vb{p}_i), \ldots, \vb*{\psi}(\vb{p}_{i + B - 1}))$.
These steps are repeated $\lceil N / B \rceil$ times\footnote{The same global feature is obtained as in the one-shot case $\vb*{\phi}(\mathcal{P}) = \max(\vb*{\psi}(\vb{p}_1), \ldots, \vb*{\psi}(\vb{p}_N))$.}.
In this way, each convolution layer only uses a buffer of size $B \times n$ ($n = 64, 128, 1024$) to store its outputs.
The pointwise feature extraction $\vb*{\psi}(\cdot)$ is parallelizable for multiple points, as its operation is independent for each point.
Our design applies a dataflow optimization to overlap the execution of different layers and exploit inter-layer parallelism.
This hides the data transfer overhead between external memory and the module as well.

Note that the original PointNet~\cite{CharlesRQi17} utilizes T-Net branches to transform the input $\mathcal{P}$ into a canonical pose and generate pose-invariant features.
T-Net is placed in the middle of two convolution layers; it takes $N$ point features $\vb{\Psi} \in \mathbb{R}^{N \times n}$ from the previous layer, predicts an affine transformation $\vb{T} \in \mathbb{R}^{n \times n}$, and passes the transformed features $\vb{\Psi} \vb{T}$ onto the next layer.
T-Net requires a buffer of size $N \times n$ ($n = 3, 64$) for $\vb{\Psi}$.
Since the registration assumes pose-sensitive global features (i.e., $\vb*{\phi}(\vb{G}_1 \cdot \mathcal{P}) \neq \vb*{\phi}(\vb{G}_2 \cdot \mathcal{P})$ holds if $\vb{G}_1 \neq \vb{G}_2$) unlike classification and segmentation tasks, T-Net branches are removed from the PointNet as in \cite{YasuhiroAoki19,DominikBauer21}.
This simplifies the network architecture and allows fully-pipelined feature extraction via dataflow optimization.
The memory consumption for layer outputs is reduced from $O(N)$ to $O(B)$ and becomes independent of input size.

\begin{figure}[h]
  \centering
  \includegraphics[keepaspectratio, width=0.6\linewidth]{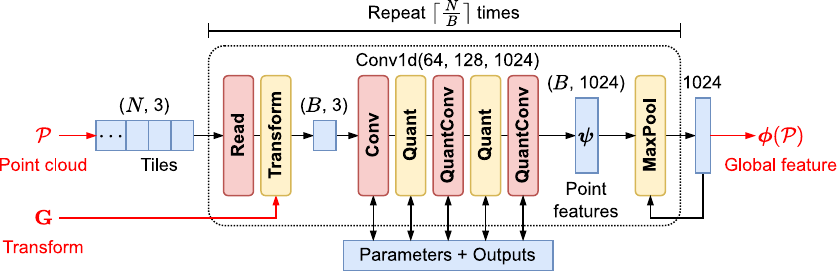}
  \caption{Block diagram of the point cloud feature extractor.}
  \label{fig:module-diagram-feature-extractor}
\end{figure}

To further save the memory consumption, the convolution layers are quantized with LLT~\cite{LongguangWang22} except the first one.
As a result of the simplified network and quantization, all parameters and intermediate results can be stored on-chip, thereby eliminating most of the off-chip data transfer.
As shown in Fig. \ref{fig:module-diagram-feature-extractor}, the module consists of four types of submodules: (\textbf{Quant})\textbf{Conv}, \textbf{Quant}, and \textbf{MaxPool}, which are described in the following.

\subsubsection{QuantConv Submodule: LLT-Quantized Convolution} \label{sec:design-quant-conv}
LLT uses two lookup tables $\mathbb{Q}_a, \mathbb{Q}_w$ for quantizing the layer inputs and weights.
The quantization process is outlined as follows (refer to \cite{LongguangWang22} for more details such as training methodologies).
The input $a \in \mathbb{R}$ and weight $w \in \mathbb{R}$ are first scaled and clipped to the range $[0, 1]$ and $[-1, 1]$, respectively:
\begin{equation}
  \hat{a} = \mathrm{clip}(a / s_a) \in [0, 1], \quad
  \hat{w} = \mathrm{clip}(w / s_w) \in [-1, 1],
  \label{eq:llt-scale-and-clip}
\end{equation}
where $s_a, s_w > 0$ are the learned scale parameters and $\mathrm{clip}(\cdot)$ is a clipping function.
Then, $\hat{a}$ and $\hat{w}$ are quantized via lookup tables and rescaled:
\begin{equation}
  \bar{a} = s_a \cdot \left( \mathbb{Q}_a(\hat{a}) / Q_a \right) \in \mathbb{R}, \quad
  \bar{w} = s_w \cdot \left( \mathbb{Q}_w(\hat{w}) / Q_w \right) \in \mathbb{R}.
  \label{eq:llt-quantize}
\end{equation}
$\mathbb{Q}_a$ ($\mathbb{Q}_w$) maps the full-precision input $\hat{a}$ ($\hat{w}$) to a $b_a$-bit ($b_w$-bit) integer in the range of $[0, Q_a]$ ($[-Q_w, Q_w]$), where $Q_a = 2^{b_a} - 1$ and $Q_w = 2^{b_w - 1} - 1$.
LLT models the quantizing function $\mathbb{Q}$ as a concatenation of step functions, with each discretized by a set of $K$ values and represented as a binary sub-table.
$K$ determines the granularity of lookup tables and is set to 9~\cite{LongguangWang22}.
Accordingly, $\mathbb{Q}_a$ ($\mathbb{Q}_w$) is of length $K Q_a + 1$ ($2K Q_w + 1$) and is formed by $Q_a$ ($2Q_w$) sub-tables\footnote{
The $i$-th sub-table in $\mathbb{Q}_a$ maps the input $\hat{a} \in [i / Q_a, (i + 1) / Q_a]$ to either $i$ or $i + 1$.
Similarly, the $i$-th sub-table in $\mathbb{Q}_w$ stores the mapping between $\hat{w} \in [(i - Q_w) / Q_w, (i - Q_w + 1) / Q_w]$ and $\{ i - Q_w, i - Q_w + 1 \}$.}.
Table indexes for the quantized values $\mathbb{Q}_a(\hat{a}), \mathbb{Q}_w(\hat{w})$ are obtained as:
\begin{equation}
  \hat{a} \mapsto \mathrm{round}(K Q_a \cdot \hat{a}) \in [0, KQ_a], \quad
  \hat{w} \mapsto \mathrm{round}(K Q_w \cdot (\hat{w} + 1)) \in [0, 2K Q_w],
  \label{eq:llt-index}
\end{equation}
where $\mathrm{round}(\cdot)$ denotes rounding to the nearest integer.

LLT can be applied to the 1D convolution in a fairly straightforward way.
Let $m, n$ denote the number of input and output channels.
The input tile $\vb{X} = \left[ x_{bj} \right] \in \mathbb{R}^{B \times m}$ and weight $\vb{W} = \left[ w_{ij} \right] \in \mathbb{R}^{n \times m}$ are quantized to produce $\bar{\vb{X}} = \left[ \bar{x}_{bj} \right]$ and $\bar{\vb{W}} = \left[ \bar{w}_{ij} \right]$.
The output $\vb{Y} = \left[ y_{bi} \right] \in \mathbb{R}^{B \times n}$ is then obtained by $y_{bi} = b_i + \sum_j \bar{x}_{bj} \bar{w}_{ij}$, where $\vb{b} = \left[ b_i \right] \in \mathbb{R}^n$ is a bias.
In this case, both operands $\bar{x}, \bar{w}$ have the same bit-widths as the original $x, w$ and the matrix-vector product is still performed in full-precision.
To address this, we rewrite the product by expanding and rearranging the terms:
\begin{align}
  y_{bi} = b_i + \textstyle \sum_{j = 1}^m \bar{x}_{bj} \bar{w}_{ij}
    &= b_i + \textstyle \sum_{j = 1}^m \left(
      s_a \cdot \mathbb{Q}_a(\hat{x}_{bj}) / Q_a \right) \cdot
      \left( s_w \cdot \mathbb{Q}_w(\hat{w}_{ij}) / Q_w \right) \label{eq:llt-conv-naive} \\
    &= b_i + s_{aw} \textstyle \sum_{j = 1}^m
      \mathbb{Q}_a(\hat{x}_{bj}) \mathbb{Q}_w(\hat{w}_{ij}), \label{eq:llt-conv}
\end{align}
where $s_{aw} = s_a s_w / (Q_a Q_w)$ is a combined scale factor.
Instead of using Eq. \ref{eq:llt-conv-naive}, i.e., performing the convolution after rescaling, \textbf{QuantConv} leverages Eq. \ref{eq:llt-conv} to perform the product between low-bit quantized integers $\mathbb{Q}_a(\hat{x}), \mathbb{Q}_w(\hat{w})$.
Except the last rescaling, most of the floating-point arithmetic is replaced by a low-bit integer arithmetic.

During inference, LLT-based quantization requires four types of parameters: a quantized weight $\mathbb{Q}_w(\hat{\vb{W}}) = \left[ \mathbb{Q}_w(\hat{w}_{ij}) \right]$ in $b_w$-bit integer format, an input lookup table $\mathbb{Q}_a$ in $b_a$-bit unsigned integer format, a bias $\vb{b}$, and a combined scale factor $s_{aw}$.
The buffer size for these parameters $N_\mathrm{QuantConv}$ is a function of the quantization bits $b_w, b_a$:
\begin{align}
  N_\mathrm{QuantConv} &= N_\mathrm{weight} + N_\mathrm{table} + N_\mathrm{bias} + N_\mathrm{scale} \nonumber \\
    &= b_w mn + b_a ((2^{b_a} - 1) K + 1) + b_v n + b_v, \label{eq:llt-conv-bufsize}
\end{align}
where $b_v$ denotes a bit-width for the non-quantized parameters and values (e.g., $\vb{b}$ and $s_{aw}$).
Compared to the standard convolution, i.e., $N_\mathrm{Conv} = N_\mathrm{weight} + N_\mathrm{bias} = b_v mn + b_v n$, the buffer size is reduced by approximately $b_v / b_w$ times when $b_a$ is small and $N_\mathrm{weight}$ is dominant.
In case of the last convolution ($m, n = 128, 1024$ and $K = 9$), $N_\mathrm{QuantConv} < N_\mathrm{Conv}$ holds if $b_w = 8, b_a \le 14$ or $b_a = 8, b_w \le 31$.
As shown in Sec. \ref{sec:eval}, 8-bit quantization (i.e., $b_w, b_a = 8$) is sufficient to achieve a reasonable accuracy, and LLT leads to memory savings in such setting.

In our design, \textbf{QuantConv} takes a quantized input $\mathbb{Q}_a(\hat{\vb{X}}) = \left[ \mathbb{Q}_a(\hat{x}_{bj}) \right]$ and computes $\vb{Z} = \mathbb{Q}_a(\hat{\vb{X}}) \mathbb{Q}_w(\hat{\vb{W}})^\top \in \mathbb{Z}^{B \times n}$.
This only requires integer arithmetic, and is easily parallelizable by unrolling the loops over the tile and output dimensions ($b, i$).
The quantization $\vb{X} \mapsto \mathbb{Q}_a(\hat{\vb{X}})$ and dequantization $\vb{Y} \gets \vb{b} + s_{aw} \vb{Z}$ are performed in the previous and next \textbf{Quant} submodules.
This saves the on-chip memory, as it reduces the input bit-width from $b_v$ to $b_a$ and the output one from $b_v$ to $b_a + b_w + \lceil \log_2 m \rceil$.

\subsubsection{Conv Submodule: 1D Convolution} \label{sec:design-conv}
\textbf{Conv} is for the standard 1D convolution.
It computes an output $\vb{Y} = \vb{X} \vb{W}^\top + \vb{b} \in \mathbb{R}^{B \times n}$ from an input $\vb{X} \in \mathbb{R}^{B \times m}$, weight $\vb{W} \in \mathbb{R}^{n \times m}$, and bias $\vb{b} \in \mathbb{R}^n$.

\subsubsection{Quant Submodule: Quantization} \label{sec:design-quant}
\textbf{Quant} serves as pre- and postprocessing steps for the LLT-based convolution and is inserted in between (\textbf{Quant})\textbf{Conv} submodules.
If its preceding layer is \textbf{QuantConv}, then it first dequantizes the input $\vb{X} \mapsto \vb{b} + s_{aw} \vb{X}$ using a bias and scale from the preceding layer.
When followed by \textbf{QuantConv}, it quantizes the output $\vb{Y} \mapsto \mathbb{Q}_a(\hat{\vb{Y}})$ using a lookup table from the next layer (Eq. \ref{eq:llt-scale-and-clip}).
The lookup operation is parallelizable by replicating the table and allowing multiple random reads.
\textbf{Quant} handles the ReLU activation and 1D batch normalization if necessary.

\subsubsection{Max Submodule: Max-pooling} \label{sec:design-max-pooling}
\textbf{MaxPool} is placed after the last \textbf{QuantConv}.
It takes a tile of pointwise features $\vb{X} = \left\{ \vb*{\psi}(\vb{p}_i), \ldots, \vb*{\psi}(\vb{p}_{i + B - 1}) \right\} \in \mathbb{R}^{B \times n}$ as input and updates the global feature $\vb*{\phi}(\mathcal{P}) \in \mathbb{R}^n$ via max-pooling, i.e., $\vb*{\phi}(\mathcal{P}) \gets \max(\vb*{\phi}(\mathcal{P}), \vb*{\psi}(\vb{p}_i), \ldots, \vb*{\psi}(\vb{p}_{i + B - 1}))$.
Similar to \textbf{Quant}, it first dequantizes the input and deals with batch normalization and ReLU if necessary.
These operations are combined into a single pipelined loop.
The design of \textbf{PointLKCore} and \textbf{ReAgentCore} is described in the following subsections.

\subsection{Case 1: Design of PointLKCore} \label{sec:design-pointlk-core}
\begin{figure}[h]
  \centering
  \includegraphics[keepaspectratio, width=0.7\linewidth]{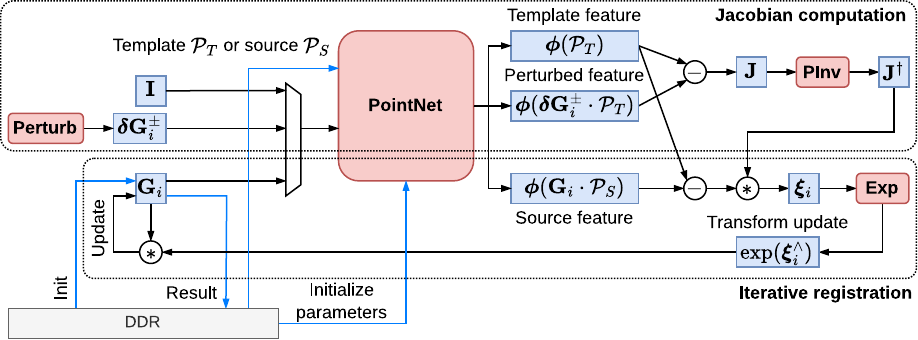}
  \caption{Block diagram of \textbf{PointLKCore}.}
  \label{fig:module-diagram-pointlk-core}
\end{figure}

\textbf{PointLKCore} is a custom accelerator core for PointNetLK.
Fig. \ref{fig:module-diagram-pointlk-core} shows the block diagram.
We first introduce an improved method for computing Jacobians to address the accuracy loss caused by quantization.

\subsubsection{Improved Jacobian Computation} \label{sec:design-pointlk-jacobian}
As described in Sec. \ref{sec:prelim-pointlk}, the Jacobian $\vb{J}$ is approximated by backward finite difference (Eq. \ref{eq:pointlk-jacobian-backward}).
Since $\vb{J}$ is involved in the solution update (Alg. \ref{alg:pointlk}, line \ref{alg:pointlk-compute-twist}) in every iteration, the quality of approximation has a major impact on the registration accuracy.
Besides, the numerical Jacobian is sensitive to quantization, as it is computed by the subtraction between two network outputs.
Considering these, \textbf{PointLKCore} uses the well-known central difference (Eq. \ref{eq:pointlk-jacobian-central}) instead of the backward one (Eq. \ref{eq:pointlk-jacobian-backward}):
\begin{align}
  \vb{J}_j^\mathrm{center} &= \frac{1}{2t_j} \left(
    \vb*{\phi}(\vb*{\delta} \vb{G}_j^- \cdot \mathcal{P}_T)
    - \vb*{\phi}(\vb*{\delta} \vb{G}_j^+ \cdot \mathcal{P}_T) \right), \quad
    (\vb*{\delta} \vb{G}_j^\pm = \exp(\pm t_j \vb{e}_j^\wedge)).
    \label{eq:pointlk-jacobian-central}
\end{align}
The backward approach has a truncation error of $O(t_j)$ ($t_j \to 0$) as shown below.
The Taylor expansion of the $k$-th element of $\vb*{\phi}(\vb*{\delta} \vb{G}_j^\pm \cdot \mathcal{P}_T)$ (up to first and second-order) yields:
\begin{align}
  \phi_k(\vb*{\delta} \vb{G}_j^\pm \cdot \mathcal{P}_T)
    \simeq \phi_k((\vb{I} \pm t_j \vb{e}_j^\wedge) \cdot \mathcal{P}_T)
    &= \phi_k(\mathcal{P}_T) \pm t_j \vb{a}_k^\top \vb{c}_j + O(t_j^2)
    \label{eq:pointlk-feature-taylor-1st} \\
    &= \phi_k(\mathcal{P}_T) \pm t_j \vb{a}_k^\top \vb{c}_j
    + \frac{1}{2} t_j^2 \vb{c}_j^\top \vb{B}_k \vb{c}_j + O(t_j^3).
    \label{eq:pointlk-feature-taylor-2nd}
\end{align}
Note that $\vb*{\delta} \vb{G}_j^\pm = \exp(\pm t_j \vb{e}_j^\wedge) \simeq \vb{I} \pm t_j \vb{e}_j^\wedge$ holds when $t_j \ll 1$.
The coefficients $\vb{a}_k \in \mathbb{R}^{3N}, \vb{B}_k \in \mathbb{R}^{3N \times 3N}, \vb{c}_j \in \mathbb{R}^{3N}$ are given by
\begin{align}
  \vb{a}_k = \left. \pdv{\phi_k(\mathcal{P})}{\mathrm{vec}(\mathcal{P})}
    \right|_{\mathcal{P} = \mathcal{P}_T}, \quad
  \vb{B}_k = \left. \pdv{\phi_k(\mathcal{P})}{\mathrm{vec}(\mathcal{P})^\top \mathrm{vec}(\mathcal{P})}
    \right|_{\mathcal{P} = \mathcal{P}_T}, \quad
  \vb{c}_j = \mathrm{vec}(\vb{e}_j^\wedge \cdot \mathcal{P}_T), \label{eq:pointlk-feature-taylor-coeffs}
\end{align}
where $\mathcal{P}_T \in \mathbb{R}^{N \times 3}$ is a template cloud of size $N$ and $\mathrm{vec}(\cdot)$ is a vectorization operator that stacks all the columns of a matrix into a single vector.
$\vb{a}$ is a direction of change in the extracted feature with respect to input point coordinates.
Substituting Eq. \ref{eq:pointlk-feature-taylor-1st} into the $k$-th element of Eq. \ref{eq:pointlk-jacobian-backward} yields
\begin{align}
  J_{jk} &= \frac{1}{t_j} \left(
    \phi_k(\mathcal{P}_T) - t_j \vb{a}_k^\top \vb{c}_j - \phi_k(\mathcal{P}_T) + O(t_j^2)
    \right) = -\vb{a}_k^\top \vb{c}_j + O(t_j).
\end{align}
This shows that backward approximation has an $O(t_j)$ error.
On the other hand, by plugging Eq. \ref{eq:pointlk-feature-taylor-2nd} into the $k$-th element of Eq. \ref{eq:pointlk-jacobian-central}, the second-order terms cancel out.
This indicates the central difference approach has a second-order accuracy $O(t_j^2)$:
\begin{align}
  J_{jk}^\mathrm{center} &= \frac{1}{2t_j}
    \left( -2t_j \vb{a}_k^\top \vb{c}_j + O(t_j^3) \right)
    = -\vb{a}_k^\top \vb{c}_j + O(t_j^2).
\end{align}
As shown in Sec. \ref{sec:eval}, the central difference gives a better accuracy especially in the quantized case.
Note that $\vb{c}_j$ is a derivative of the transformed point coordinates $\exp(-\vb*{\xi}^\wedge) \mathcal{P}_T$ with respect to the $j$-th twist parameter $\xi_j$ ($j = 1, \ldots, 6$):
\begin{align}
  \left. \pdv{\exp(-\vb*{\xi}^\wedge) \mathcal{P}_T}{\xi_j} \right|_{\vb*{\xi} = \vb{0}}
    &= \lim_{t \to 0} \left. \frac{1}{t}
      \left( \exp(-t \vb{e}_j^\wedge) \exp(-\vb*{\xi}^\wedge) \mathcal{P}_T
      - \exp(-\vb*{\xi}^\wedge) \mathcal{P}_T \right) \right|_{\vb*{\xi} = \vb{0}} \\
    &\simeq \lim_{t \to 0} \frac{1}{t}
      \left((\vb{I} - t \vb{e}_j^\wedge) \mathcal{P}_T - \mathcal{P}_T \right)
    = - \vb{e}_j^\wedge \mathcal{P}_T. \label{eq:pointlk-warp-jacobian}
\end{align}
From Eqs. \ref{eq:pointlk-feature-taylor-coeffs} and \ref{eq:pointlk-warp-jacobian}, it turns out that $-\vb{a}_k^\top \vb{c}_j$ represents the $(j, k)$ component of an analytical Jacobian proposed in \cite{XueqianLi21}, where $\vb{a}$ and $\vb{c}$ are referred to as the feature gradient and warp Jacobian, respectively.
The central approach is sufficient in terms of accuracy as shown in Fig. \ref{fig:ex1-bits-and-jacobian}.
Besides, the analytical Jacobian significantly increases the runtime (Sec. \ref{sec:eval-pointlk-jacobi}), due to the computational cost for a feature gradient of size $(N, 3, 1024)$.
The forward or backward approach requires six perturbed template features to compute $\vb{J}$, while the central approach requires twelve.
\textbf{PointLKCore} implements these three approaches for performance comparison.

\subsubsection{Registration with PointLKCore} \label{sec:design-pointlk-reg}
As shown in Fig. \ref{fig:module-diagram-pointlk-core}, \textbf{PointLKCore} contains four modules, namely \textbf{Perturb}, \textbf{PInv}, \textbf{Exp}, and \textbf{PointNet}.
At initialization, the core moves PointNet parameters including convolution weights and lookup tables from an external buffer to the relevant on-chip buffers.
It then proceeds to the registration process.

\textbf{PointLKCore} first extracts a feature $\vb*{\phi}(\mathcal{P}_T)$ of a template $\mathcal{P}_T$ and computes a numerical Jacobian $\vb{J}$ using one of the three approaches presented in Sec. \ref{sec:design-pointlk-jacobian}.
When the forward or backward difference is used, it extracts six perturbed features $\{ \vb*{\phi}(\vb*{\delta} \vb{G}_j^\pm \cdot \mathcal{P}_T) \}$ from the perturbed templates $\{ \vb*{\delta} \vb{G}_j^\pm \cdot \mathcal{P}_T \}$ ($j = 1, \ldots, 6$).
In the $j$-th iteration, \textbf{Perturb} generates an infinitesimal transform $\vb*{\delta} \vb{G}_j^\pm \in \SE(3)$ and \textbf{PointNet} produces a perturbed feature $\vb*{\phi}(\vb*{\delta} \vb{G}_j^\pm \cdot \mathcal{P}_T)$, from which the $j$-th column of the Jacobian $\vb{J}_j$ is calculated (Eq. \ref{eq:pointlk-jacobian-backward}).
If the central difference is used, \textbf{PointLKCore} extracts twelve perturbed features.
In each iteration $j \in [1, 6]$, it extracts a pair of features $(\vb*{\phi}(\vb*{\delta} \vb{G}_j^+ \cdot \mathcal{P}_T), \vb*{\phi}(\vb*{\delta} \vb{G}_j^- \cdot \mathcal{P}_T))$ and fills the $j$-th column of $\vb{J}$ (Eq. \ref{eq:pointlk-jacobian-central}).
After the Jacobian is obtained, \textbf{PInv} computes its pseudoinverse $\vb{J}^\dagger$ and \textbf{PointLKCore} moves on to the iterative registration (Alg. \ref{alg:pointlk}, lines \ref{alg:pointlk-opt-begin}-\ref{alg:pointlk-opt-end}).
The core incrementally updates the transform $\vb{G}_i \in \SE(3)$ from source to template.
In iteration $i$, it extracts a feature $\vb*{\phi}(\vb{G}_{i - 1} \cdot \mathcal{P}_S)$ of a transformed source $\vb{G}_{i - 1} \cdot \mathcal{P}_S$ and solves for the optimal twist parameters $\vb{\Delta}\vb*{\xi}_i^* \in \mathbb{R}^6$.
\textbf{Exp} computes $\exp(\vb{\Delta}\vb*{\xi}_i^{*\wedge})$, which is left-multiplied to the current transform: $\vb{G}_i \gets \exp(\vb{\Delta}\vb*{\xi}_i^{*\wedge}) \cdot \vb{G}_{i - 1}$.
This continues until convergence or the maximum number of iterations.
The result $\vb{G}$ is written to the external buffer.

\subsubsection{Perturb Module: Generate Perturbation Transforms} \label{sec:design-pointlk-perturb}
\textbf{Perturb} is to generate an infinitesimal rigid transform $\vb*{\delta} \vb{G}_j^\pm = \exp(\pm t_j \vb{e}_j^\wedge) \in \SE(3)$ ($j = 1, \ldots, 6$) which is used to perturb the template $\mathcal{P}_T$.
Note that, it simplifies to $\vb*{\delta} \vb{G}_j^\pm = \vb{I}_{4 \times 4} \pm t_j \vb{e}_j^\wedge$ and can be written down explicitly\footnote{Each element in $\vb*{\delta} \vb{G}_j$ is $0$, $1$, or $\pm t_j$.}, because $t_j$ is small and $\vb{e}_j \in \mathbb{R}^6$ is a one-hot vector.
If $1 \le j \le 3$, the upper-left $3 \times 3$ block of $\vb*{\delta} \vb{G}_j^\pm$ represents a small rotation of $\pm t_j$ radians around the $x, y, z$ axis.
In case of $4 \le j \le 6$, the rightmost column of $\vb*{\delta} \vb{G}_j^\pm$ represents a translation by $\pm t_j$ along the $x, y, z$ axis.

\subsubsection{PInv Module: Pseudoinverse} \label{sec:design-pointlk-pinv}
\textbf{PInv} computes a pseudoinverse $\vb{J}^\dagger \in \mathbb{R}^{6 \times 1024} = (\vb{J}^\top \vb{J})^{-1} \vb{J}^\top$ of the Jacobian $\vb{J}$, which involves the inversion of a $6 \times 6$ symmetric matrix $\vb{J}^\top \vb{J}$.
Since it is small, \textbf{PInv} adopts a simple approach for the inversion.
It partitions $\vb{J}^\top \vb{J}$ into four submatrices of size $3 \times 3$, inverts each submatrix using an adjoint method, and applies a blockwise inversion formula (assuming that block diagonals are invertible) to obtain $(\vb{J}^\top \vb{J})^{-1}$.

\subsubsection{Exp Module: Exponential Map} \label{sec:design-pointlk-exp}
\textbf{Exp} deals with $\exp(\cdot): \se(3) \to \SE(3)$ (refer to \cite{Barfoot17} for details).
It takes a 6D twist parameter $\vb*{\xi} = \left( \vb*{\omega}, \vb*{\rho} \right)$ and computes a rigid transform $\vb{G} = \left[ \vb{R} \mid \vb{t} \right]$, where $\vb*{\omega}, \vb*{\rho} \in \mathbb{R}^3$ are the rotational and translational components.
The rotation $\vb{R} = \exp(\vb*{\omega}^\wedge)$ is obtained by the famous Rodrigues' formula, while the translation is written as $\vb{t} = \vb{J}_l(\vb*{\omega}) \vb*{\rho}$ ($\vb{J}_l(\vb*{\omega})$ is a left-Jacobian of $\SO(3)$).

\subsubsection{PointNet Module: Feature Extraction} \label{sec:design-pointlk-pointnet}
\textbf{PointNet} is for feature extraction as explained in Sec. \ref{sec:design-feature-extractor}.
Each convolution is followed by batch normalization and ReLU.
It takes a point cloud $\mathcal{P} \in \mathbb{R}^{N \times 3}$ along with a rigid transform $\vb{G} \in \SE(3)$ to compute $\vb*{\phi}(\vb{G} \cdot \mathcal{P})$.
The input $\mathcal{P}$ is first transformed by $\vb{G}$ before being passed to a stack of layer submodules.
The pipeline stages thus include (i) the input data transfer from an external memory, (ii) rigid transform, and (iii) layer submodules.
Each stage is parallelized via array partitioning and loop unrolling.

\subsection{Case 2: Design of ReAgentCore} \label{sec:design-reagent-core}
\begin{figure}[h]
  \centering
  \includegraphics[keepaspectratio, width=0.8\linewidth]{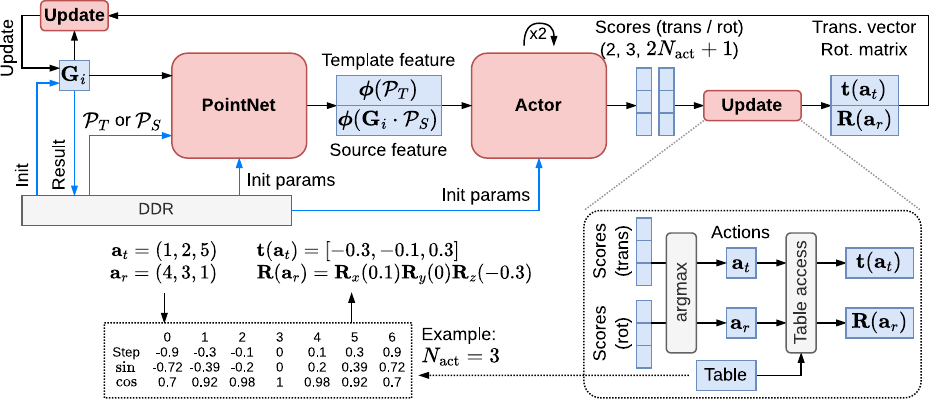}
  \caption{Block diagram of \textbf{ReAgentCore}.}
  \label{fig:module-diagram-reagent-core}
\end{figure}

\textbf{ReAgentCore} integrates the PointNet feature extractor (Sec. \ref{sec:design-feature-extractor}) and the other components for actor networks and rigid transform.
Fig. \ref{fig:module-diagram-reagent-core} depicts the block diagram.

\subsubsection{Registration with ReAgentCore} \label{sec:design-reagent-reg}
\textbf{ReAgentCore} consists of three modules: \textbf{PointNet}, \textbf{Actor}, and \textbf{Update}.
At initialization, it transfers the parameters for PointNet and two actor networks from an external memory to the on-chip buffers.
It then extracts a template feature $\vb*{\phi}(\mathcal{P}_T)$ using \textbf{PointNet} and proceeds to iteratively update $\vb{G}$.
In iteration $i$, it computes a feature $\vb*{\phi}(\vb{G}_{i - 1} \cdot \mathcal{P}_S)$ of the transformed source.
The concatenated feature vector $(\vb*{\phi}(\vb{G}_{i - 1} \cdot \mathcal{P}_S), \vb*{\phi}(\mathcal{P}_T))$ is fed to the \textbf{Actor} two times to determine the translational and rotational actions $\vb{a}_t, \vb{a}_r$, which are incorporated into $\vb{G}_{i - 1}$ by \textbf{Update} module to obtain a new transform $\vb{G}_i$.
After $I_{\max}$ iterations, the result $\vb{G}$ is written back to the external buffer.

\subsubsection{Update Module: Transform Update} \label{sec:design-reagent-update}
In this module, the outputs from actor networks (of size $(2, 3, 2N_\mathrm{act} + 1)$) are converted to the action vectors $\vb{a}_t, \vb{a}_r$ using a table $\mathbb{T}$, from which a new transform $\vb{G}_i$ is obtained (Sec. \ref{sec:prelim-reagent}).
$\mathbb{T}$ is of size $(2N_\mathrm{act} + 1) \times 3$, and stores a mapping from $2N_\mathrm{act} + 1$ action labels to the corresponding step sizes (Eq. \ref{eq:reagent-step-sizes}) as well as their cosine and sine values\footnote{We store cosine and sine values to avoid the trigonometric operations for converting Euler angles to rotation matrices.}.

\subsubsection{Actor Module: Action Decision} \label{sec:design-reagent-actor}
\begin{figure}[h]
  \centering
  \includegraphics[keepaspectratio, width=0.7\linewidth]{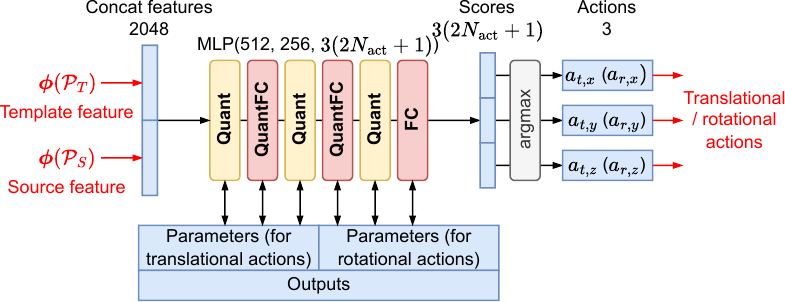}
  \caption{Block diagram of the Actor module in \textbf{ReAgentCore}.}
  \label{fig:module-diagram-actor-critic}
\end{figure}

Given the current state $(\vb*{\phi}(\vb{G}_{i - 1} \cdot \mathcal{P}_S), \vb*{\phi}(\mathcal{P}_T)) \in \mathbb{R}^{2048}$, \textbf{Actor} decides the action that reduces the alignment error between two point clouds.
It implements an actor network consisting of three fully-connected (FC) layers of size $(512, 256, 3(2N_\mathrm{act} + 1))$, each followed by ReLU activation.
Similar to the feature extractor (Sec. \ref{sec:design-feature-extractor}), weight parameters in the FC layers are quantized by LLT except the last one to save on-chip memory.
As shown in Fig. \ref{fig:module-diagram-actor-critic}, \textbf{Actor} contains a set of layer submodules, (\textbf{Quant})\textbf{Conv} and \textbf{Quant}, and two sets of parameter buffers (for translation and rotation).

\subsubsection{PointNet Module: Feature Extraction} \label{sec:design-reagent-pointnet}
\textbf{PointNet} encodes the point cloud $\vb{G} \cdot \mathcal{P} \in \mathbb{R}^{N \times 3}$ into a latent feature $\vb*{\phi}(\vb{G} \cdot \mathcal{P})$.
A submodule is added before the first convolution layer to transform $\mathcal{P}$ with $\vb{G} = \left[ \vb{R} \mid \vb{t} \right] \in \SE(3)$ in a disentangled manner (Sec. \ref{sec:prelim-reagent}).
The dataflow optimization is applied such that the input data transfer, rigid transform, and layer submodules form a single pipeline.

\subsection{Details and Board-level Implementation} \label{sec:board-level-impl}
\begin{figure}[h]
  \centering
  \includegraphics[keepaspectratio, width=0.45\linewidth]{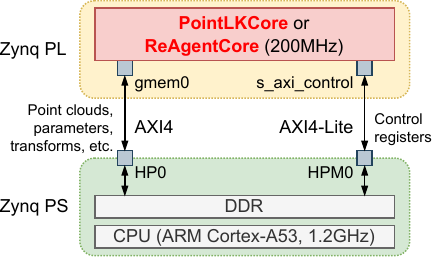}
  \caption{Board-level implementation for the Xilinx Zynq SoC.}
  \label{fig:board-level-design}
\end{figure}

Fig. \ref{fig:board-level-design} shows a board-level implementation of the proposed core for Xilinx Zynq SoC.
The core has a 128-bit AXI manager port to transfer the necessary data (e.g., point clouds, network parameters, and transforms) in bursts, which is directly connected to a high-performance subordinate port (HP0).
The core uses a 32-bit AXI-Lite subordinate port as well, which allows the host program to access the control registers and configure the algorithmic parameters (e.g., the number of iterations $I_{\max}$ and the step size $t_i$ for Jacobian computation) through the high-performance manager port (HPM0).
The operation frequency of the core is set to 200MHz throughout this paper.

In quantized layers, inputs $\mathbb{Q}(\hat{\vb{X}}) \in \mathbb{R}^{B \times m}$ and weights $\mathbb{Q}(\hat{\vb{W}}) \in \mathbb{R}^{n \times m}$ are $b_a$-bit signed and $b_w$-bit unsigned integers, respectively.
We set $b_a = b_w$ throughout the evaluation as in \cite{LongguangWang22}.
The output bit-width is adjusted to store the matrix product $\mathbb{Q}(\hat{\vb{X}}) \mathbb{Q}(\hat{\vb{W}})^\top$ with no precision loss (i.e., $b_a + b_w + \lceil \log_2 m \rceil$ bits).
The parameters and outputs for the rest non-quantized layers (e.g., batch normalization) are 32-bit fixed-point with 16.16 format (16-bit fraction and 16-bit integer part).
We use 32-bit floating-point for other mathematical operations (e.g., pose composition, exponential map, pseudoinverse, etc.) to prevent numerical instabilities.

%% file: dse.tex

\section{Design Space Exploration} \label{sec:dse}
The proposed cores comprise a set of submodules, with each having its own design parameters (i.e., loop unrolling factors).
It is thus intractable to run the synthesis for each design point in the exponentially growing design space.
This section describes the performance and resource modeling for the proposed cores to reduce the cost for design space exploration (DSE) and quickly find optimal design points under resource constraints.

\subsection{Modeling the Point Cloud Feature Extractor} \label{sec:model-feature-extractor}
We first derive the number of operations (\#OPs) $\mathrm{OP}$, clock cycle latency $C$, amount of data transfer $D$ (bytes), and resource usage $R_x$ ($x \in \{ \text{DSP}, \text{BRAM}, \text{URAM} \}$).
As for the PointNet feature extraction (Fig. \ref{fig:module-diagram-feature-extractor}), they are modeled as (superscripted P):
\begin{align}
  \left\{ \begin{array}{l}
    \mathrm{OP}^\mathrm{P}
      = N \cdot \left( \sum_s \mathrm{OP}^{\mathrm{P}, s} \right) \\
    C^\mathrm{P} = \left( \left\lceil \frac{N}{B} \right\rceil - 1 \right)
      \max_s C^{\mathrm{P}, s}
      + \sum_s C^{\mathrm{P}, s} \\
    D^\mathrm{P} = 16N \\
    R_x^\mathrm{P} = \sum_s R_x^{\mathrm{P}, s}
      \quad (x \in \{ \text{DSP}, \text{BRAM} \}),
  \end{array} \right.
  \label{eq:model-feature-extractor}
\end{align}
where $B, N$ are the tile size (Sec. \ref{sec:design-feature-extractor}) and number of points.
$\mathrm{OP}^\mathrm{P}$ and $R_x^\mathrm{P}$ are the sums of the \#OPs and resource usage for each pipeline stage $s$.
The data transfer size is simply $D^\mathrm{P} = 16N$ (bytes) because network parameters are stored on-chip and a point cloud is transferred as $N$ 128-bit packets with each containing three 32-bit floating-point coordinates.
The overall latency $C^\mathrm{P}$ is obtained by the pipeline stage with the largest latency $\max_s C^{\mathrm{P}, s}$, number of tiles $\lceil N / B \rceil$, and pipeline latency $\sum_s C^{\mathrm{P}, s}$.
The computation in each stage can be parallelized by unrolling the loops over the (i) points in a tile and (ii) output dimensions; thus, each stage $s$ has unrolling factors $P^{\mathrm{P}, s, p}, P^{\mathrm{P}, s, o}$ as design variables.
$C^{\mathrm{P}, s}$ and $R_x^{\mathrm{P}, s}$ are the linear functions of these factors as well.
For instance, the model for \textbf{QuantConv}($m, n$) is given as:
\begin{align}
  \left\{ \begin{array}{l}
    \mathrm{OP}^{\mathrm{P}, s} = 2Bmn \\
    C^{\mathrm{P}, s}(P^{\mathrm{P}, s, p}, P^{\mathrm{P}, s, o})
      = \left\lceil \frac{B}{P^{\mathrm{P}, s, p}} \right\rceil
      \left\lceil \frac{n}{P^{\mathrm{P}, s, o}} \right\rceil
      \left( \mathrm{II}_\mathrm{loop} (m - 1) + C_\mathrm{loop} \right) \\
    R_\mathrm{DSP}^{\mathrm{P}, s}(P^{\mathrm{P}, s, p}, P^{\mathrm{P}, s, o})
      = \eta P^{\mathrm{P}, s, p} P^{\mathrm{P}, s, o} \\
    R_\mathrm{BRAM}^{\mathrm{P}, s}(P^{\mathrm{P}, s, o})
      = R_\mathrm{BRAM}(mn, b_w, \frac{P^{\mathrm{P}, s, o}}{2}),
  \end{array} \right.
\end{align}
where $\mathrm{II}_\mathrm{loop}, C_\mathrm{loop}$ are the iteration interval (II) and latency of the loop over the input dimension $m$.
$\mathrm{OP}^{\mathrm{P}, s}$ comes from a matrix multiplication between a quantized input $\mathbb{Q}(\hat{\vb{X}}) \in \mathbb{Z}^{B \times m}$ and a quantized weight $\mathbb{Q}(\hat{\vb{W}}) \in \mathbb{Z}^{n \times m}$.
$R_\mathrm{DSP}^{\mathrm{P}, s}$ increases linearly with the unrolling factors, where $\eta$ is a DSP cost per PE ($\eta = 1, 3$ for \textbf{QuantConv} and \textbf{Conv} according to the HLS report).
BRAM usage is due to the quantized weight $\mathbb{Q}(\hat{\vb{W}})$.
One BRAM block has a capacity of 18Kb with the maximum bit-width of 36; BRAM usage for a buffer ($w$-bit, length $s$, partition factor $P$) is modeled as in \cite{JasonCong17}:
\begin{align}
  R_\mathrm{BRAM}(s, w, P)
    = P \left\lceil \frac{sw}{P \left\lceil \frac{w}{36} \right\rceil \cdot 18\text{Kb}} \right\rceil
    \left\lceil \frac{w}{36} \right\rceil.
  \label{eq:bram-model}
\end{align}
The partitioning factors are set as half of the unrolling factors since BRAMs are dual-port.
URAM utilization is modeled by Eq. \ref{eq:bram-model} as well, except that the capacity and the maximum bit-width are doubled.
An input lookup table $\mathbb{Q}_a$ is duplicated according to the unrolling factors to fully parallelize the quantization process, and its BRAM utilization is given as:
\begin{align}
  R_\mathrm{BRAM}(P^{\mathrm{P}, s, p} P^{\mathrm{P}, s, o}
    \mathrm{LUTSize}(b_a, K), b_a, P^{\mathrm{P}, s, p} P^{\mathrm{P}, s, o}),
\end{align}
where $\mathrm{LUTSize}(b_a, K) = K(2^{b_a} - 1) + 1$ is a table size (Sec. \ref{sec:design-quant-conv}).

To reduce the complexity of DSE, we only consider unrolling factors of a stage $s^*$ with the largest \#OPs, \textbf{QuantConv}(128, 1024), as design variables.
For another stage $s \neq s^*$, we adjust the unrolling factors to balance the stage latencies:
\begin{align}
  P^{\mathrm{P}, s, p} = \min(B, \frac{C^{\mathrm{P}, s}(1, 1)}{
    C^{\mathrm{P}, s^*}(P^{\mathrm{P}, s^*, p}, P^{\mathrm{P}, s^*, o})}), \quad
  P^{\mathrm{P}, s, o} = \max(1, \frac{C^{\mathrm{P}, s}(B, 1)}{
    C^{\mathrm{P}, s^*}(P^{\mathrm{P}, s^*, p}, P^{\mathrm{P}, s^*, o})}).
\end{align}
The loop over points is unrolled first, and then over the output dimensions\footnote{We assume that $P^{\mathrm{P}, s, p}$ and $P^{\mathrm{P}, s, o}$ are factors of the tile size $B$ and the output dimensions $n$, respectively.}.
The unrolling and partitioning factors gradually increase in the later convolution layers due to the increasing number of output channels from 64 to 1024.
The layer outputs are smaller than the parameters and hence are implemented using the distributed RAM, as larger partitioning factors for parallel reads would cause the under-utilization of the BRAM capacity.
The number of free design parameters for feature extractor are thus reduced to three: two unrolling factors $P^{\mathrm{P}, s^*, p}, P^{\mathrm{P}, s^*, o}$ for the longest stage and a tile size $B$.

\subsection{Modeling the PointLKCore} \label{sec:model-pointlk-core}
For \textbf{PointLKCore}, the performance and resource model are given in Eq. \ref{eq:model-pointlk-core} (superscripted L):
\begin{align}
  \left\{ \begin{array}{l}
    \mathrm{OP}^{\mathrm{L}} = (I_\mathrm{Jacobi} + 1) \mathrm{OP}^{\mathrm{P}}
      + \mathrm{OP}^{\mathrm{L}, \mathrm{PInv}}
      + I_{\max} (\mathrm{OP}^{\mathrm{P}} + \mathrm{OP}^{\mathrm{L}, \mathrm{Update}}) \\
    C^{\mathrm{L}} = (I_\mathrm{Jacobi} + 1) C^{\mathrm{P}}
      + C^{\mathrm{L}, \mathrm{PInv}}
      + I_{\max} (C^{\mathrm{P}} + C^{\mathrm{L}, \mathrm{Update}}) \\
    D^{\mathrm{L}} = (I_\mathrm{Jacobi} + I_{\max} + 1) D^{\mathrm{P}}
      + (I_{\max} + 1) D^\mathrm{Trans} \\
    R_\mathrm{DSP}^{\mathrm{L}} = R_\mathrm{DSP}^{\mathrm{P}}
      + R_\mathrm{DSP}^{\mathrm{L}, \mathrm{PInv}} + R_\mathrm{DSP}^{\mathrm{L}, \mathrm{Update}} \\
    R_\mathrm{BRAM}^{\mathrm{L}} = R_\mathrm{BRAM}^{\mathrm{P}}
      + R_\mathrm{BRAM}^{\mathrm{L}, \mathrm{Feature}}
      + R_\mathrm{BRAM}^{\mathrm{L}, \mathrm{Jacobi}} + R_\mathrm{BRAM}^{\mathrm{L}, \mathrm{PInv}}.
  \end{array} \right.
  \label{eq:model-pointlk-core}
\end{align}
$I_\mathrm{Jacobi} = 6, 12$ is a number of perturbed features to compute Jacobians, and $I_{\max}$ is the maximum iterations.
$\mathrm{OP}^\mathrm{L}$ consists of the three terms: (i) \#OPs to extract a template feature $\vb*{\phi}(\mathcal{P}_T)$ as well as perturbed features $\{ \vb*{\phi}(\vb*{\delta} \vb{G}_i^\pm \cdot \mathcal{P}_T) \}$, (ii) \#OPs for pseudoinverse $\vb{J}^\dagger$, and (iii) \#OPs for iterative registration with each iteration involving the extraction of a source feature $\vb*{\phi}(\vb{G}_{i - 1} \cdot \mathcal{P}_S)$ and a transform update $\vb{G}_i \gets \exp(\vb{\Delta}\vb*{\xi}_i^\wedge) \cdot \mathbf{G}_{i - 1}$.
The latency $C^\mathrm{L}$ and DSP usage $R_\mathrm{DSP}^\mathrm{L}$ are defined in a similar way.
$D^\mathrm{L}$ is determined by the number of PointNet runs (i.e., point cloud size), an initial transform $\vb{G}_0$, and output transforms $\{ \vb{G}_1, \ldots, \vb{G}_{I_{\max}} \}$; $D^\mathrm{Trans} = 48$ (bytes) is a size of a $3 \times 4$ rigid transform.
$R_\mathrm{BRAM}^\mathrm{L}$ is a sum of BRAM blocks for the feature extractor, output features, Jacobian matrix $\vb{J}$, and pseudoinverse $\vb{J}^\dagger$.
The unrolling factors for the last PointNet pipeline stage determines $R_\mathrm{BRAM}^{\mathrm{L}, \mathrm{Feature}}$.

As expected, the terms for feature extraction (with a superscript P) are dominant in Eq. \ref{eq:model-pointlk-core}; for instance, we observe $C^\mathrm{P}$ are 108.5x/6.9x larger than $C^{\mathrm{L}, \mathrm{Update}}$/$C^{\mathrm{L}, \mathrm{PInv}}$ in our design.
While the pseudoinverse and transform update are also parallelizable by loop unrolling, the unrolling factors are fixed and excluded from the design variables.
The relevant terms (e.g., $\mathrm{C}^{\mathrm{L}, \mathrm{PInv}}$ and $R_\mathrm{DSP}^{\mathrm{L}, \mathrm{Update}}$) are hence treated as constants and obtained by running HLS for once.
\textbf{PointLKCore} has the same set of design parameters ($P^{\mathrm{P}, s^*, p}, P^{\mathrm{P}, s^*, o}, B$) as in Sec. \ref{sec:model-feature-extractor}.

\subsection{Modeling the ReAgentCore} \label{sec:model-reagent-core}
\textbf{ReAgentCore} is modeled by the dominant terms (for PointNet and two actor networks) as in Eq. \ref{eq:model-reagent-core} (superscripted R):
\begin{align}
  \left\{ \begin{array}{ll}
    \mathrm{OP}^{\mathrm{R}} = \mathrm{OP}^\mathrm{P}
      + I_{\max} (\mathrm{OP}^\mathrm{P} + 2 \mathrm{OP}^{\mathrm{R}, \mathrm{Actor}})
    & \mathrm{OP}^{\mathrm{R}, \mathrm{Actor}} = \sum_l \mathrm{OP}^{\mathrm{R}, \mathrm{Actor}, l} \\
    C^{\mathrm{R}} = C^\mathrm{P}
      + I_{\max} (C^\mathrm{P} + 2 C^{\mathrm{R}, \mathrm{Actor}})
    & C^{\mathrm{R}, \mathrm{Actor}} = \sum_l C^{\mathrm{R}, \mathrm{Actor}, l} \\
    D^{\mathrm{R}} = (I_{\max} + 1) D^\mathrm{P}
      + (I_{\max} + 1) D^\mathrm{Trans} \\
    R_\mathrm{DSP}^{\mathrm{R}} = R_\mathrm{DSP}^{\mathrm{P}}
      + R_\mathrm{DSP}^{\mathrm{R}, \mathrm{Actor}}
    & R_\mathrm{DSP}^{\mathrm{R}, \mathrm{Actor}}
      = \sum_l R_\mathrm{DSP}^{\mathrm{R}, \mathrm{Actor}, l} \\
    R_\mathrm{BRAM}^{\mathrm{R}} = R_\mathrm{BRAM}^{\mathrm{P}}
      + 2R_\mathrm{BRAM}^{\mathrm{R}, \mathrm{Actor}}
      + R_\mathrm{BRAM}^{\mathrm{R}, \mathrm{Feature}}
    & R_\mathrm{BRAM}^{\mathrm{R}, \mathrm{Actor}}
      = \sum_l R_\mathrm{BRAM}^{\mathrm{R}, \mathrm{Actor}, l}.
  \end{array} \right.
  \label{eq:model-reagent-core}
\end{align}
$\mathrm{OP}^\mathrm{R}$ and $C^\mathrm{R}$ are based on that \textbf{ReAgentCore} first extracts a template feature and then repeats the feature extraction and action decision alternately.
The data transfer size $D^\mathrm{R}$ is similar to $D^\mathrm{L}$ in Sec. \ref{sec:model-pointlk-core}.
$R_\mathrm{BRAM}^\mathrm{R}$ consists of the BRAMs for the feature extractor, two actor networks, and output features.
$\mathrm{OP}^{\mathrm{R}, \mathrm{Actor}}$, $C^{\mathrm{R}, \mathrm{Actor}}$, and $R_x^{\mathrm{R}, \mathrm{Actor}}$ are the sums of the \#OPs, latencies, and resource usages for all layer submodules in Fig. \ref{fig:module-diagram-reagent-core}.
Similar to Sec. \ref{sec:model-feature-extractor}, the computation can be parallelized by applying the loop unrolling on the output dimension; an unrolling factor $P^{\mathrm{R}, l, o}$ should be set for each layer $l$.
For simplicity, a single factor $P^{\mathrm{R}, l^*, o}$ is chosen for the largest layer $l^*$, \textbf{Quant}(2048, 512), and factors $\{ P^{\mathrm{R}, l, o} \}$ for the other layers $l \neq l^*$ are automatically determined via the latency ratio:
\begin{align}
  P^{\mathrm{R}, l, o} = \max(1, \frac{C^{\mathrm{R}, l}(1)}{C^{\mathrm{R}, l^*}(P^{\mathrm{R}, l^*, o})}).
\end{align}
\textbf{ReAgentCore} thus has four design parameters in total: $(P^{\mathrm{P}, s^*, p}, P^{\mathrm{P}, s^*, o}, B)$ (for feature extractor) and $P^{\mathrm{R}, l^*, o}$ (for actor).
\textbf{ReAgentCore} makes use of URAMs to store the quantized weight $\mathbb{Q}(\hat{\vb{W}})$ for the largest FC layer \textbf{QuantFC}(2048, 512) to prevent an over-utilization of BRAMs.
In addition, the layer outputs are implemented using LUTRAMs to avoid an inefficient BRAM usage (Sec. \ref{sec:model-feature-extractor}).

\subsection{DSE Method} \label{sec:dse-method}
There are several choices of performance metrics to use as an exploration objective.
One approach is to use the following~\cite{ChenZhang15,StefanoRibes20}:
\begin{equation}
  \mathrm{Perf} = \min(\mathrm{CP}, \mathrm{CTC} \cdot \mathrm{BW}_{\max}), \quad
  \mathrm{CP} = \frac{\mathrm{OP}}{C \cdot \frac{1}{f}}, \quad
  \mathrm{CTC} = \frac{\mathrm{OP}}{D}
  \label{eq:model-roofline}
\end{equation}
where $f$, $\mathrm{CP}$, $\mathrm{CTC}$, and $\mathrm{BW}_{\max}$ denote the operating frequency of the IP core (Hz), computational performance (ops/s), computation-to-communication (CTC) ratio (ops/bytes, i.e., \#OPs per byte of data moved from/to the off-chip memory), and maximum off-chip bandwidth (bytes/s).
In our board-level design (Fig. \ref{fig:board-level-design}), $f = 200\text{MHz}$ and $\mathrm{BW}_{\max}$ is computed as $200\text{MHz} \cdot 128\text{bit} = 3.2\text{GB/s}$, assuming that 128-bit data is transferred every clock cycle~\cite{AlecLu22}.

Eq. \ref{eq:model-roofline} suggests the attainable performance, $\mathrm{Perf}$ (ops/s), is bounded by either the amount of computing resources available on FPGA (first term) or the off-chip memory bandwidth (second term).
In our cases, $\mathrm{CP}$ is far lower than $\mathrm{CTC} \cdot \mathrm{BW}_{\max}$, indicating that the proposed cores are compute-bound rather than memory-bound.
For instance, we observe 140.8x and 187.1x differences between the first and second terms for \textbf{PointLKCore} and \textbf{ReAgentCore} with the final design parameters ($\mathrm{CP} = 404.8, 280.6\text{Gops/s}$, $\mathrm{CTC} \cdot \mathrm{BW}_{\max} = 5.7 \cdot 10^4, 5.25 \cdot 10^4\text{Gops/s}$), respectively.
Since all network parameters fit within the on-chip memory, the data transfer size $D$ is significantly reduced and only the point clouds and rigid transforms are transferred from/to off-chip during registration\footnote{Since BRAM is not fully utilized, the whole point cloud can be stored on-chip as well (if $N$ is relatively small), which would further increase the CTC ratio.}.
This leads to the high CTC ratio ($\mathrm{CTC} = 1.78 \cdot 10^4, 1.64 \cdot 10^4\text{ops/bytes}$) and pushes the design points towards the compute-bound region.

We therefore use the overall latency $C$ as a simple performance metric; the objective of DSE is to find a set of design parameters that minimize the latency $C$ while satisfying the resource constraints (i.e., $R_x$ should not exceed the configured threshold).
Since there are only three or four design variables and the design space is relatively small (contains around 1M design points), a simple brute-force search is feasible.
Tables \ref{tbl:design-params} presents the resulting design parameters.

\begin{table}[htbp]
  \centering
  \caption{Design parameters for \textbf{PointLKCore} and \textbf{ReAgentCore}}
  \label{tbl:design-params}
  \begin{tabularx}{0.9\linewidth}{Y|cccccccc} \hline
    \multicolumn{9}{c}{Design parameters for \textbf{PointLKCore}} \\ \hline
    \multirow{3}{*}{$B$}
      & \multicolumn{8}{c}{$(P^{\mathrm{P}, s, p}, P^{\mathrm{P}, s, o})$ for PointNet pipeline stages} \\ \cline{2-9}
    & Read & Transform & \textbf{Conv} & \textbf{Quant}
      & \textbf{QuantConv} & \textbf{Quant}
      & \textbf{QuantConv} & \textbf{MaxPool} \\
    & & & (3, 64) & 64 & (64, 128) & 128 & (128, 1024) & 1024 \\ \hline
    2 & N/A & (1, 1) & (2, 4) & (1, 1) & (2, 64) & (1, 1) & (2, 512) & (1, 8) \\ \hline \hline
    \multicolumn{9}{c}{Design parameters for \textbf{ReAgentCore}} \\ \hline
    \multirow{3}{*}{$B$}
      & \multicolumn{8}{c}{$(P^{\mathrm{P}, s, p}, P^{\mathrm{P}, s, o})$ for PointNet pipeline stages} \\ \cline{2-9}
    & Read & Transform & \textbf{Conv} & \textbf{Quant}
      & \textbf{QuantConv} & \textbf{Quant}
      & \textbf{QuantConv} & \textbf{MaxPool} \\
    & & & (3, 64) & 64 & (64, 128) & 128 & (128, 1024) & 1024 \\ \hline
    14 & N/A & (2, 1) & (7, 1) & (1, 1) & (14, 8) & (1, 1) & (14, 64) & (1, 8) \\ \hline
  \end{tabularx}
  \begin{tabularx}{0.9\linewidth}{YYYYYY}
    \multicolumn{6}{c}{$P^{\mathrm{R}, l, o}$ for actor network layers} \\ \hline
    \textbf{Quant} & \textbf{QuantFC} & \textbf{Quant}
      & \textbf{QuantFC} & \textbf{Quant} & \textbf{FC} \\
    2048 & (2048, 512) & 512 & (512, 256) & 256 & 33 \\ \hline
    1 & 128 & 1 & 32 & 1 & 2 \\ \hline
  \end{tabularx}
\end{table}

%% file: eval.tex

\section{Evaluation} \label{sec:eval}
In this section, we evaluate the performance of the proposed cores (Sec. \ref{sec:design}) in comparison with existing registration methods.

\subsection{Experimental Setup} \label{sec:eval-setup}
We develop \textbf{PointLKCore} and \textbf{ReAgentCore} in HLS C++, which contains a set of HLS preprocessor directives for design optimizations (e.g., loop unrolling and array partitioning).
We run Vitis HLS 2022.1 to generate the IP core, and then Vivado 2022.1 to synthesize the board-level design (Fig. \ref{fig:board-level-design}).
Xilinx ZCU104 is chosen as an embedded FPGA platform, which integrates a quad-core ARM Cortex-A53 CPU (1.2GHz), an FPGA chip (XCZU7EV-2FFVC1156), and a 2GB DRAM on the same board.
The board runs the Ubuntu 20.04-based Pynq Linux 2.7 OS, which provides a Python API to interact with the accelerator kernels on the PL side.
The host programs are written in Python with PyTorch 1.10.2 and Open3D 0.15.1\footnote{We compile PyTorch 1.10.2 from source with \texttt{-O3} optimization and auto-vectorization (ARM Neon SIMD instructions) enabled using GCC 9.3.0.}.
For performance comparison, we use a desktop computer and two Nvidia embedded GPUs (Jetson Xavier/Nano) as well, which are summarized in Table \ref{tbl:machines}.
As baselines, we use the published code of PointNetLK~\cite{YasuhiroAoki19} and ReAgent~\cite{DominikBauer21}, to which we add a Pynq-based code to run the proposed cores.
We use the implementation of PointNetLK-v2~\cite{XueqianLi21} to compute analytical Jacobian matrices.
In addition, we consider two well-known classical methods: ICP (point-to-point and point-to-plane) and FGR (Fast Global Registration), both of which are available in Open3D.

\begin{table}[htbp]
  \centering
  \caption{Machine Specifications} \label{tbl:machines}
  \begin{tabular}{l|lll} \hline
    & Desktop & NVidia Jetson Xavier NX & NVidia Jetson Nano \\ \hline
    \multirow{2}{*}{CPU} & Intel Xeon W-2235
      & Nvidia Carmel ARM v8.2 & ARM Cortex-A57 \\
    & (6C/12T, 3.8GHz) & (6C/6T, 1.4GHz) & (4C/4T, 1.43GHz) \\
    DRAM & 64GB & 8GB & 4GB \\
    GPU & Nvidia GeForce RTX 3090
      & 384-core Nvidia Volta & 128-core Nvidia Maxwell \\
    \multirow{2}{*}{OS Image} & \multirow{2}{*}{Ubuntu 20.04.6}
      & Nvidia JetPack 5.1 & Nvidia JetPack 4.6.3 \\
    & & (Ubuntu 20.04.6) & (Ubuntu 18.04.6) \\
    Python & 3.10.12 & 3.8.2 & 3.6.15 \\
    Open3D & 0.17.0 & 0.15.1 & 0.15.1 \\
    PyTorch & 2.0.1 (CUDA 11.7) & 2.0.0+nv23.05 (CUDA 11.4) & 1.10.0 (CUDA 10.2) \\ \hline
  \end{tabular}
\end{table}

\subsubsection{Model Training} \label{sec:eval-model-training}
For PointNetLK, we first train the full-precision (i.e., FP32) model for 100 epochs with a learning rate of $10^{-3}$ on the noisy point clouds (Sec. \ref{sec:eval-dataset}), and the model parameters are used to initialize the LLT-quantized model.
We finetune the quantized model for another 100 epochs with a learning rate of $10^{-4}$.
The learning rate is decayed by a factor of 0.8 after every 10 epochs.
The batch size is set to 32, the step size $t_i$ to 0.01, the maximum number of iterations $I_{\max}$ to 10, and the convergence threshold $\varepsilon$ to $10^{-7}$ (Alg. \ref{alg:pointlk}, line \ref{alg:pointlk-check-convergence}).
Adam is used as an optimizer with default settings of $\beta_1 = 0.9$ and $\beta_2 = 0.999$.
Following \cite{XiaoshuiHuang20}, we jointly train PointNetLK with a classifier or decoder.
The classifier is a three-layer MLP (1024, 512, 256, $N_c$), with each layer followed by batch normalization and ReLU except the last one.
$N_c$ is a total number of object categories in the dataset (e.g., 40 for ModelNet40).
Similarly, the decoder is a stack of three fully-connected layers of size (1024, 512, 256, $3N$); the first two are followed by batch normalization and ReLU, whereas the last one is followed by tanh activation.
It produces 3D point coordinates (in the range of $[-1, 1]$) for $N$ points to reconstruct the input point cloud.

PointNetLK is trained to minimize the registration error $\mathcal{L}_\mathrm{pose}(\vb{G}) = \left\| \vb{G}^{-1} \vb{G}^* - \vb{I} \right\|_2^2$ between estimated and ground-truth rigid transforms $\vb{G}, \vb{G}^* \in \SE(3)$.
We use a feature alignment error $\mathcal{L}_\mathrm{feat}$ as well (Eq. \ref{eq:pointlk-objective-naive}).
For the classifier and decoder, we use a cross-entropy $\mathcal{L}_\mathrm{cls}$ and a reconstruction error $\mathcal{L}_\mathrm{dec}$, respectively; the latter is defined by a Chamfer distance between $\mathcal{P}_T$ and $\vb{G} \cdot \mathcal{P}_S$ (see \cite{XiaoshuiHuang20}).
The final loss function $\mathcal{L}$ for PointNetLK is a weighted sum of these losses:
\begin{align}
  \mathcal{L} = \textstyle \sum_{x \in \left\{ \text{pose}, \text{feat}, \text{cls}, \text{dec} \right\}}
    \lambda_x \mathcal{L}_x,
\end{align}
where $\lambda_x$ is a hyperparameter weight for each term ($\lambda_\mathrm{cls}, \lambda_\mathrm{dec} = 0$ if the classifier or decoder is not used).
We set $\lambda_\mathrm{pose}$ to 100 and the rest to 1.

For ReAgent, we take the three steps for training: (i) we first pre-train the full-precision model on the noise-free point clouds for 100 epochs as in \cite{DominikBauer21}, then (ii) train the same model on the noisy point clouds for additional 50 epochs.
We initialize the quantized model with the full-precision one and (iii) finetune it for 50 epochs.
The learning rate is set to $10^{-3}$ for (i) and $10^{-4}$ for (ii)--(iii).
The optimizer, learning rate scheduler, and batch size are the same as in PointNetLK.
Only IL is used for training; RL is not employed as it does not seem to improve the accuracy in our case.
Given a training sample $(\vb{G} \cdot \mathcal{P}_S, \mathcal{P}_T, \vb{a}_t^*, \vb{a}_r^*)$, where $\vb{a}_t^*, \vb{a}_r^*$ are the translational and rotational actions of the expert and $\vb{G}$ is a randomly-generated rigid transform (Sec. \ref{sec:eval-dataset}), the model is trained to align a source $\vb{G} \cdot \mathcal{P}_S$ with a template $\mathcal{P}_T$ by choosing the same actions.
The standard cross-entropy is used as a loss function, such that model mimics the expert demonstration.

\subsubsection{Datasets} \label{sec:eval-dataset}
Following \cite{DominikBauer21}, we employ two representative point cloud datasets: ModelNet40~\cite{ZhirongWu15} and ScanObjectNN~\cite{MikaelaAngelinaUy19}.
ModelNet40 contains a total of 12,311 synthetic CAD models from 40 object categories (9,843 for training and 2,468 for testing).
We use the preprocessed dataset provided by the authors of \cite{CharlesRQi17}, which contains point clouds extracted by uniformly sampling 2,048 points from the model surface.
The point clouds are zero-centered and scaled to fit within a unit sphere.
We use the first 20 categories (\textbf{Seen}: airplane to lamp) for both training and evaluation, while the remaining 20 categories (\textbf{Unseen}; laptop to xbox) are only used for evaluation to validate the generalization to objects unseen during training.
ScanObjectNN is a set of segmented objects extracted from real-world indoor scenes.
The test split contains 581 samples from 15 object categories, with each sample having 2,048 points.

For each sample $\mathcal{P}$ in the dataset, we obtain an input $(\mathcal{P}_S, \mathcal{P}_T, \vb{G}^*$) as follows: from $\mathcal{P}$, we subsample $N$ points independently to create a pair of source and template\footnote{There is no exact one-to-one correspondence between a source and template because they are independently sampled.}.
We then generate a rigid transform $\vb{G}$ from a random Euler angle within $[0^\circ, \theta_{\max}]$ and a random translation within $[-t_{\max}, t_{\max}]$ on each axis.
The transform is applied to the source, such that $\vb{P}_T = \vb{G}^* \cdot \vb{P}_S$ and $\vb{G}^* = \vb{G}^{-1}$.
For ModelNet40, we jitter the points in $\mathcal{P}_S$ and $\mathcal{P}_T$ independently using a random Gaussian noise, which is sampled from $\mathcal{N}(0, r_\mathrm{std})$ and clipped to $[-r_\mathrm{clip}, r_\mathrm{clip}]$.
On the other hand, we do not add a noise to the point clouds in ScanObjectNN, as they are acquired from real-world RGB-D scans and are already affected by sensor noise.
Unless otherwise noted, $I_{\max}$ is set to 20 for PointNetLK (10 for ReAgent), $N$ to 1024 for ModelNet40 (2048 for ScanObjectNN), $(\theta_{\max}, t_{\max})$ to $(45^\circ, 0.5)$, and $(r_\mathrm{std}, r_\mathrm{clip})$ to $(0.01, 0.05)$.
Point-to-plane ICP and FGR requires point normals\footnote{FGR extracts an FPFH feature for each point, which requires normal information.}.
Since normals are not available in ScanObjectNN, we additionally perform $k$NN search and PCA (Principal Component Analysis) to estimate them.

\subsection{Registration Accuracy} \label{sec:eval-accuracy}
We first evaluate the registration accuracy of the proposed cores in comparison with baselines.
Following \cite{ZiJianYew20,DominikBauer21}, we compute ISO (Isotropic error) and CD (Chamfer Distance) as error metrics.

Table \ref{tbl:accuracy-m40-scnn} (left two columns) compares the accuracy on ModelNet40.
\textbf{IP} refers to the registration with the proposed cores.
PointNetLK is jointly trained with a decoder and uses a central difference for Jacobian computation.
As seen in rows 1--2, PointNetLK with 8-bit quantization ($b_w = b_a = 8$) achieves a comparable accuracy to the FP32 model, with a $0.259^\circ$ increase in the ISO error for Unseen set, showing that LLT can be applied to a geometric task as well as semantic tasks.
In ReAgent (rows 4--5), the ISO error increases by $0.626^\circ$ and $0.474^\circ$ for Seen and Unseen sets, respectively, with 8-bit quantization.
While PointNetLK uses a DNN only for feature extraction, ReAgent employs two actor networks as wells, and is more likely to be affected by quantization due to the increased number of model parameters.
Both cores maintain the same level of accuracy compared to their software counterparts (rows 2--3, 5--6).
The 8-bit PointNetLK and ReAgent outperform two ICP variants and FGR in terms of CD, indicating that two point clouds are more closely aligned after the registration is complete.
Point-to-point (pt2pt) ICP suffers from the lack of one-to-one point correspondences between a source and template (Sec. \ref{sec:eval-dataset}), which is often the case in practice, and point-to-plane (pt2pl) ICP is still prone to wrong or missing correspondences.
This highlights the benefit of correspondence-free approaches that operate on the global features of point clouds.
FGR gives on-par accuracy with the proposed cores, while it requires surface normals as well.
Both PointNetLK and ReAgent have similar registration errors for Seen and Unseen sets, which confirms that they generalize well to unseen object categories.

Table \ref{tbl:accuracy-m40-scnn} (the rightmost column) shows the results for ScanObjectNN.
Notably, while PointNetLK and ReAgent are trained on the synthetic CAD dataset, they handle real-world point clouds consisting of incomplete and partial objects due to occlusions, without degrading the accuracy as in ICP.
Point-to-plane ICP performs poorly because it uses unreliable surface normals estimated from noisy points.

\begin{table}[htbp]
  \centering
  \caption{Accuracy on the ModelNet40 and ScanObjectNN datasets ($\downarrow$ indicates that lower is better)}
  \label{tbl:accuracy-m40-scnn}
  \begin{tabular}{c|cc|rrr|rrr|rrr} \hline
    \multirow{2}{*}{} & \multirow{2}{*}{$b$} & \multirow{2}{*}{IP} &
      \multicolumn{3}{c|}{ModelNet40 (Seen)} &
      \multicolumn{3}{c|}{ModelNet40 (Unseen)} &
      \multicolumn{3}{c}{ScanObjectNN} \\
    & & & \multicolumn{2}{c}{ISO ($\downarrow$)} & CD ($\downarrow$) &
      \multicolumn{2}{c}{ISO ($\downarrow$)} & CD ($\downarrow$) &
      \multicolumn{2}{c}{ISO ($\downarrow$)} & CD ($\downarrow$) \\ \hline
    & & & R & t & $\times 10^{-3}$ &
      R & t & $\times 10^{-3}$ & R & t & $\times 10^{-3}$ \\ \hline
    \multirow{3}{*}{PointNetLK} & FP32 & & 2.556 & 0.0211 & 0.807 &
      1.921 & 0.0174 & 0.910 & 1.376 & 0.0104 & 0.533 \\
    & 8 & & 2.568 & 0.0194 & 0.790 &
      2.180 & 0.0170 & 0.861 & 1.462 & 0.0115 & 0.610 \\
    & 8 & $\checkmark$ & 2.499 & 0.0189 & 0.783 &
      2.157 & 0.0169 & 0.881 & 1.574 & 0.0117 & 0.660 \\ \hline
    \multirow{3}{*}{ReAgent} & FP32 & & 3.088 & 0.0249 & 0.745 &
      2.501 & 0.0199 & 0.963 & 1.416 & 0.0132 & 0.503 \\
    & 8 & & 3.714 & 0.0283 & 0.773 &
      2.975 & 0.0231 & 0.981 & 3.050 & 0.0245 & 0.927 \\
    & 8 & $\checkmark$ & 3.786 & 0.0290 & 0.778 &
      2.969 & 0.0234 & 0.968 & 2.987 & 0.0243 & 0.925 \\ \hline
    ICP (pt2pt) & FP32 & & 9.861 & 0.0855 & 4.905 &
      10.004 & 0.0787 & 4.617 & 13.640 & 0.104 & 5.091 \\
    ICP (pt2pl) & FP32 & & 7.308 & 0.0572 & 5.003 &
      7.845 & 0.0559 & 5.168 & 18.509 & 0.136 & 19.870 \\
    FGR & FP32 & & 3.877 & 0.0318 & 1.470 &
      2.973 & 0.0243 & 1.528 & 2.884 & 0.0232 & 1.725 \\ \hline
  \end{tabular}
\end{table}

\begin{figure}[htbp]
  \centering
  \includegraphics[keepaspectratio, width=\linewidth]{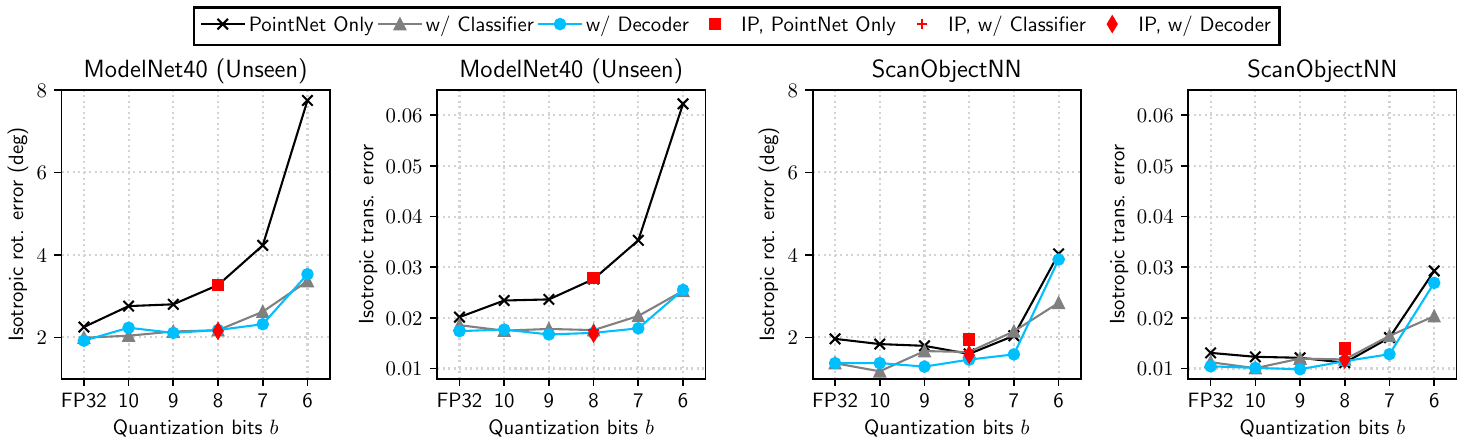}
  \caption{Accuracy of PointNetLK under different training methodologies and quantization bits.
  PointNetLK uses a central difference for Jacobian approximation.}
  \label{fig:ex1-bits-and-training}
\end{figure}

\begin{figure}[htbp]
  \centering
  \includegraphics[keepaspectratio, width=\linewidth]{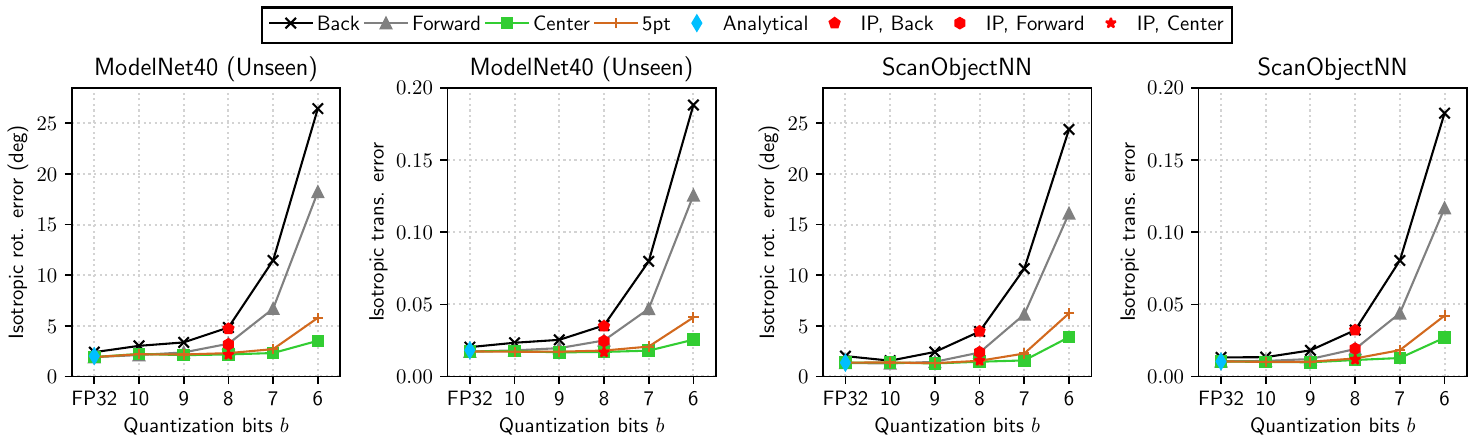}
  \caption{Accuracy of PointNetLK under different Jacobian computation methods and quantization bits.
  PointNetLK is trained with a decoder.}
  \label{fig:ex1-bits-and-jacobian}
\end{figure}

Figs. \ref{fig:ex1-bits-and-training}--\ref{fig:ex1-bits-and-jacobian} plot the ISO error of PointNetLK under different number of quantization bits $b$.
Fig. \ref{fig:ex1-bits-and-training} shows that the accuracy improves when a classifier or decoder is jointly trained, especially in case of lower bits (e.g., 6).
The vanilla PointNetLK is trained with a feature alignment error (Eq. \ref{eq:pointlk-objective-naive}) to guide PointNet to extract similar features for well-aligned point clouds.
In this case, since the objective is to minimize a difference between two features, the feature itself may not capture the geometric structure of the point cloud.
The result indicates that extracting a distinctive feature, which is transferable to other tasks (e.g., classification and reconstruction), is important in the feature-based registration.
For Unseen set (ModelNet40), the ISO error of 6-bit PointNetLK is $7.74^\circ$, which is brought down to $3.36^\circ$ and $3.53^\circ$ with a classifier and decoder ($4.03^\circ$, $2.84^\circ$, and $3.89^\circ$ for ScanObjectNN).
The unsupervised training with a decoder still only requires raw point clouds as input, and is more beneficial than using a classifier, considering that correct labels may not be available or a single point cloud can contain multiple objects.
As shown in Fig. \ref{fig:ex1-bits-and-training}, the accuracy drops sharply when $b < 7$ in both datasets, showing that $b = 8$ gives the best compromise between resource utilization and accuracy.
The proposed cores (marked with red) maintain the quality of results as their software counterparts.

\begin{figure}[htbp]
  \centering
  \includegraphics[keepaspectratio, width=\linewidth]{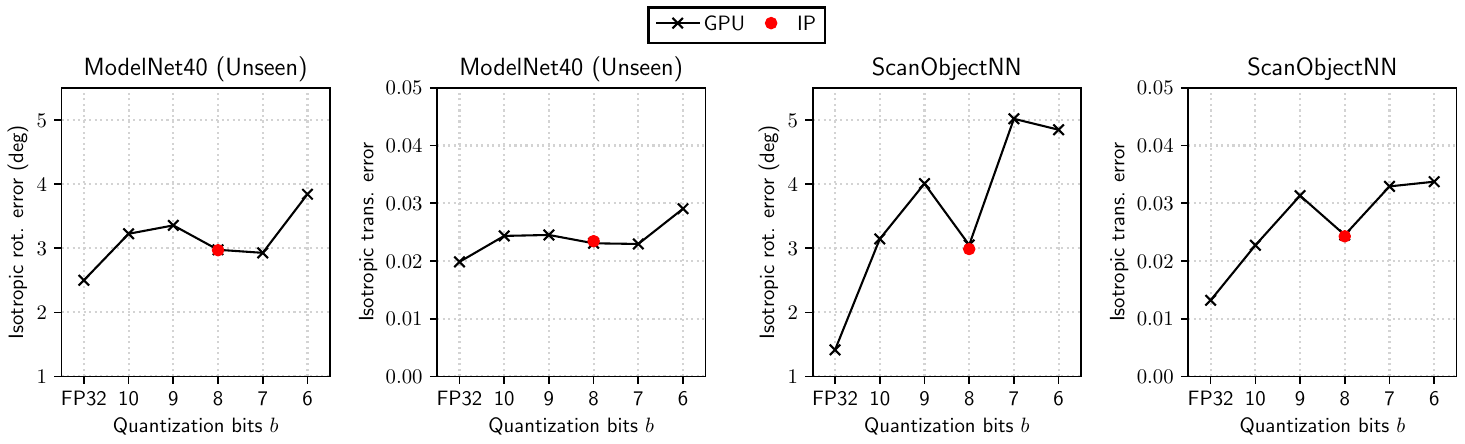}
  \caption{Accuracy of ReAgent under different quantization bits.}
  \label{fig:ex1-reagent-bits}
\end{figure}

Fig. \ref{fig:ex1-bits-and-jacobian} highlights the benefit of using central difference approximation for Jacobians.
When the forward or backward difference is used, the accuracy significantly degrades with $b \le 7$ due to the first-order truncation error $O(t_i)$.
Fig. \ref{fig:ex1-bits-and-jacobian} includes the results for a five-point method as well.
While it has a smaller error of $O(t_i^4)$, it does not provide better accuracy and is more sensitive to noise ($b = 6$) compared to the central difference (with an $O(t_i^2)$ error).
Thus, the central difference gives the best trade-off between computational cost and approximation accuracy.
Fig. \ref{fig:ex1-reagent-bits} shows the ISO error of ReAgent for a varying $b$.
$b = 8$ achieves the on-par or even better accuracy than $b = 9, 10$, suggesting that $b = 8$ is sufficient for both datasets.

\begin{figure}[htbp]
  \centering
  \includegraphics[keepaspectratio, width=\linewidth]{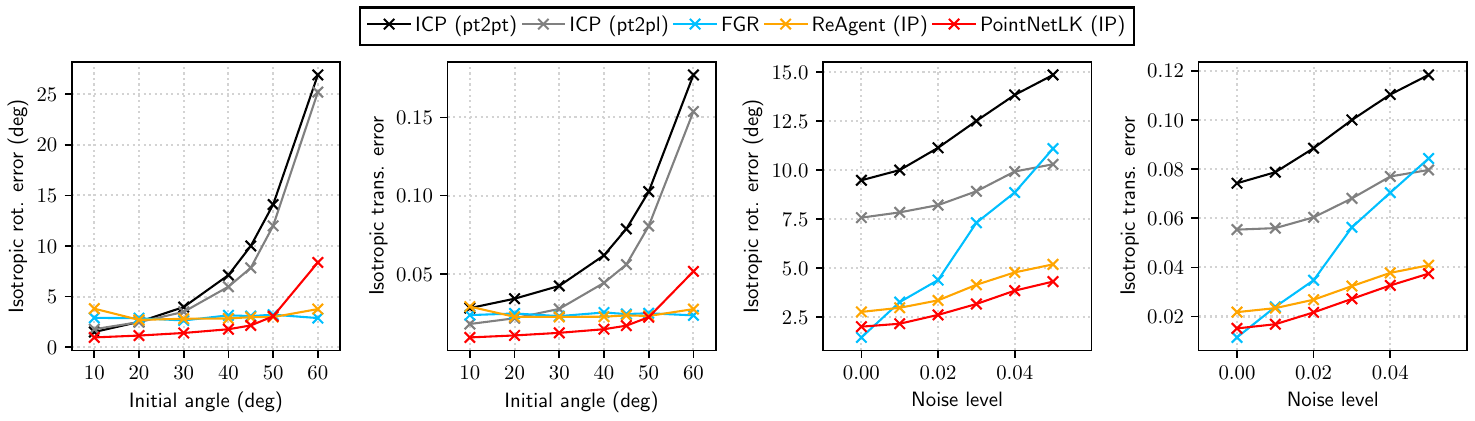}
  \caption{Accuracy of registration methods (ModelNet40, Unseen).}
  \label{fig:ex2-methods-m40}
\end{figure}

\begin{figure}[htbp]
  \centering
  \includegraphics[keepaspectratio, width=\linewidth]{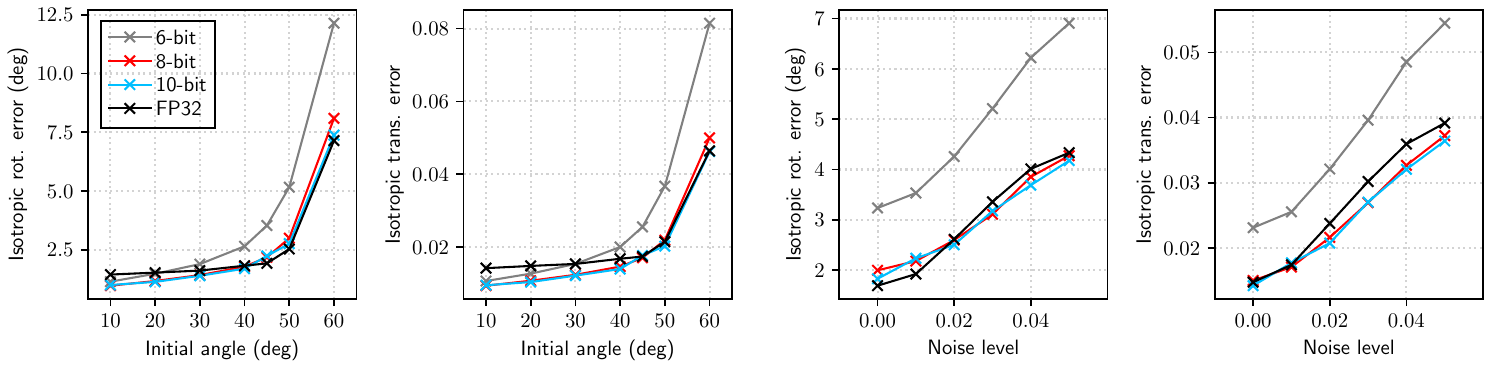}
  \caption{Accuracy of PointNetLK for different quantization bits (ModelNet40, Unseen).}
  \label{fig:ex2-quant-pointlk-m40}
\end{figure}

\begin{figure}[htbp]
  \centering
  \includegraphics[keepaspectratio, width=\linewidth]{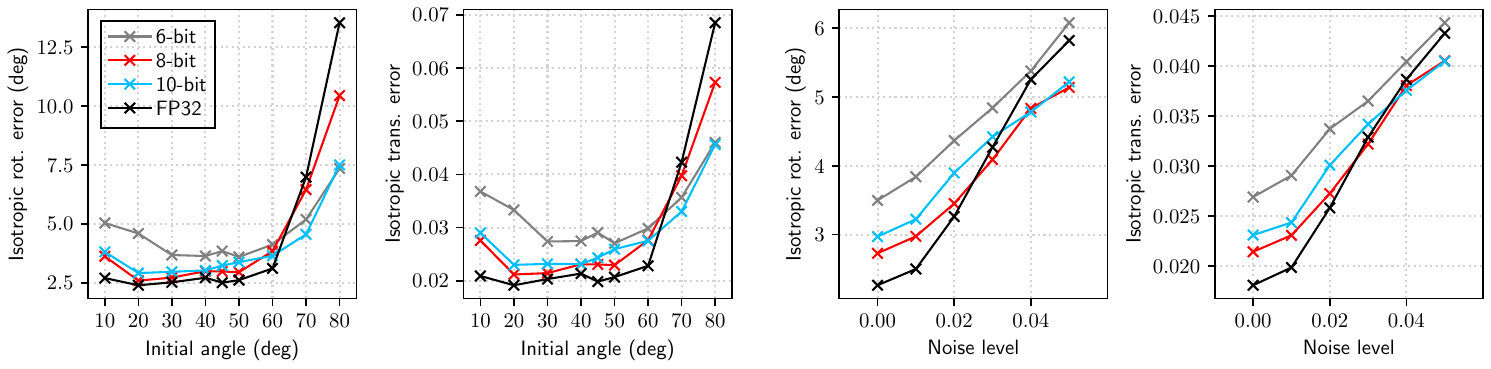}
  \caption{Accuracy of ReAgent for different quantization bits (ModelNet40, Unseen).}
  \label{fig:ex2-quant-reagent-m40}
\end{figure}

In Figs. \ref{fig:ex2-methods-m40}--\ref{fig:ex2-quant-reagent-m40}, the ISO error is evaluated with varying initial rotation angles $\theta_{\max} \in [10, 60]$ ($t_{\max} = 0.5$, $(r_\mathrm{std}, r_\mathrm{clip}) = (0.01, 0.05)$) or with varying noise levels $r_\mathrm{std} \in [0.01, 0.05]$ ($(\theta_{\max}, t_{\max}) = (45^\circ, 0.5)$, $r_\mathrm{clip} = 0.1$).
As shown in Fig. \ref{fig:ex2-methods-m40}, both ICP variants perform poorly with a larger $\theta_{\max}$, because they rely on the nearest neighbor search and fail to find correct correspondences in the two point clouds.
While point-to-plane ICP is less sensitive to noise than the point-to-point variant, it still gives a larger error than the other methods.
FGR is a global registration method and is able to handle large initial displacements.
Despite that noise is not applied to point normals, the accuracy of FGR drops sharply with the increasing $r_\mathrm{std}$, as FPFH feature is computed based on the local geometry around a point, which is corrupted by the noise.
PointNetLK and ReAgent are more robust to initial misalignment and noise, and outperform the classical methods, showing the advantage of deep features over simple geometric features (e.g., normals) or handcrafted ones.
PointNetLK achieves slightly better accuracy than ReAgent on a wide range of $\theta_{\max}$ and $r_\mathrm{std}$, as ReAgent uses the discrete set of actions and PointNetLK is less affected by quantization.

As seen in Fig. \ref{fig:ex2-quant-pointlk-m40}, 8-bit or 10-bit PointNetLK has a comparable accuracy to FP32.
The 6-bit one is prone to noise and suffers from the considerable accuracy drop unless $\theta_{\max} \le 30^\circ$.
PointNetLK is unable to converge in case of $\theta_{\max} \ge 60^\circ$, as it is a local method and assumes that input point clouds are roughly aligned.
It successfully registers when $\theta_{\max}$ is within the range of $[0^\circ, 50^\circ]$; note that $\theta_{\max}$ is set to $45^\circ$ during training.
Similarly, 8-bit is sufficient to retain the accuracy of ReAgent (Fig. \ref{fig:ex2-quant-reagent-m40}), and the 6-bit version gives a higher error even for a smaller $\theta_{\max}$.
The ISO error starts to grow rapidly when $\theta_{\max} \ge 70^\circ$ and otherwise remains less than $5^\circ$ (and 0.03).
Considering that ReAgent is trained with $\theta_{\max} = 45^\circ$, it generalizes to more difficult settings with larger initial misalignments.

\subsection{Computation Time} \label{sec:eval-time}
\begin{figure}[htbp]
  \begin{minipage}[b]{0.48\linewidth}
    \centering
    \includegraphics[keepaspectratio, height=1.7in]{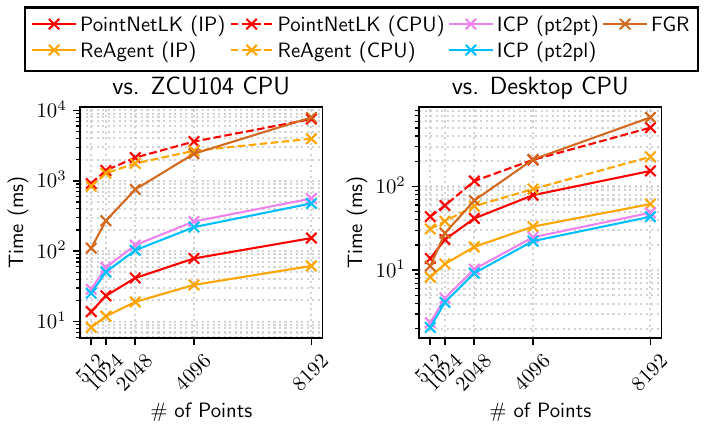}
    \subcaption{Different methods}
    \label{fig:ex3-methods-m40-table}
  \end{minipage}
  \begin{minipage}[b]{0.48\linewidth}
    \centering
    \includegraphics[keepaspectratio, height=1.7in]{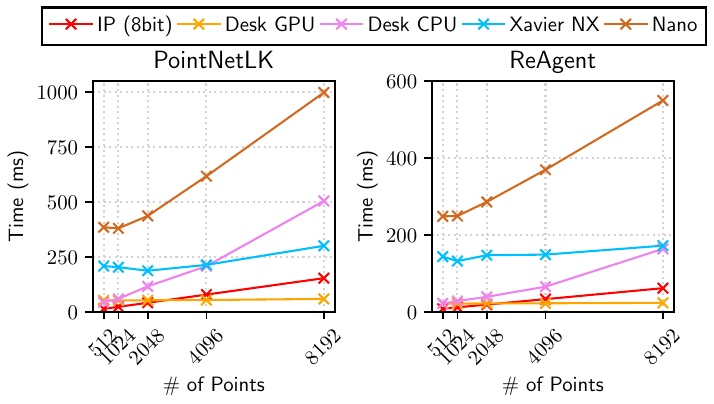}
    \subcaption{Different platforms}
    \label{fig:ex3-machines-m40-table}
  \end{minipage}
  \caption{Comparison of the computation time (ModelNet40, Table).}
  \label{fig:ex3-methods-machines-m40-table}
\end{figure}

Fig. \ref{fig:ex3-methods-m40-table} shows the execution times of the proposed cores in comparison with baselines.
For evaluation, we use ModelNet40 table category containing 100 samples, and vary the input size $N$ from 512 to 8,192.
On ZCU104, the cores run the fastest among all baselines on a wide range of $N$, and greatly improve the trade-off between accuracy and running time.
In case of $N = 4096$, \textbf{PointLKCore} and \textbf{ReAgentCore} achieve a 45.75x and 44.08x speedup over their software counterparts, leading to (3.35x, 2.79x, 30.71x) and (7.95x, 6.62x, 72.93x) faster registration than point-to-point ICP, point-to-plane ICP, and FGR.
Notably, the performance gain improves with the larger input size: for $N = 16384$, they provide a (3.98x, 3.43x, 78.16x) and (10.13x, 8.71x, 198.64x) speedup than these baselines.
The proposed cores even outperform the software counterparts and FGR running on the desktop CPU (dashed lines).
While two ICP variants run faster than our methods, they are not accurate enough and sensitive to noise as shown in Fig. \ref{fig:ex2-methods-m40}.
FGR is sensitive to noise as well, and takes 2.67x and 6.33x longer than \textbf{PointLKCore} and \textbf{ReAgentCore}, respectively.

The execution time of PointNetLK and ReAgent shows a linear increase with $N$, reflecting the $O(N)$ computational complexity of PointNet, whereas that of FGR grows faster than $O(N)$.
This is because FGR involves $k$NN search for every point to extract FPFH features, which amounts to at least $O(N \log N)$ complexity.
Fig. \ref{fig:ex3-machines-m40-table} plots the execution time of PointNetLK and ReAgent on various platforms (Table \ref{tbl:machines}).
While the desktop GPU surpasses our cores when $N \ge 4096$, our cores are consistently faster than the desktop CPU and embedded GPUs.
For $N = 4096$, PointNetLK and ReAgent are (2.64x, 7.83x, 2.71x) and (1.98x, 11.13x, 4.49x) faster on the FPGA than on the desktop CPU, Jetson Nano, and Xavier NX, respectively.

\begin{figure}[htbp]
  \begin{minipage}[b]{0.48\linewidth}
    \centering
    \includegraphics[keepaspectratio, height=1.1in]{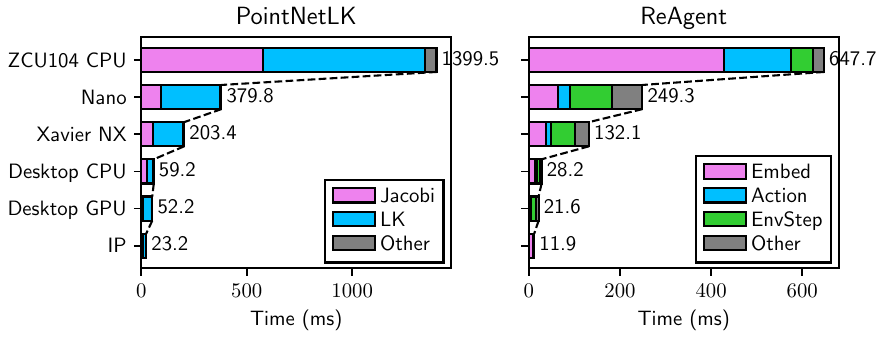}
    \caption{Computation time breakdown.}
    \label{fig:ex3-break-m40-table}
  \end{minipage}
  \begin{minipage}[b]{0.48\linewidth}
    \centering
    \includegraphics[keepaspectratio, height=1.1in]{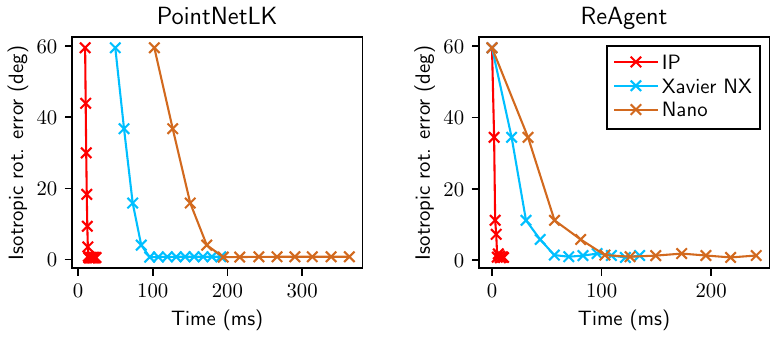}
    \caption{Evolution of the rotational ISO error.}
    \label{fig:ex5-m40-table}
  \end{minipage}
\end{figure}

Fig. \ref{fig:ex3-break-m40-table} shows the execution time breakdown of PointNetLK and ReAgent on various platforms (Table \ref{tbl:machines}).
We use ModelNet40 table category and set to $N = 1024$.
For the proposed cores, we obtain the time breakdowns based on the clock cycle information from HLS reports.
In PointNetLK, Jacobian computation (Jacobi) and iterative registration (LK) are the two major steps and dominate the execution time.
\textbf{PointLKCore} speeds up both processes by (60.89x, 55.71x) (from (576.71ms, 766.38ms) to (9.47ms, 13.76ms)), yielding an overall speedup of 60.25x.
In case of ReAgent on ZCU104, PointNet feature embedding (Embed) takes up 66.12\% of the execution time, and other steps such as the actor network (Action) and state update (EnvStep) account for a non-negligible portion as well.
By implementing the entire registration flow on FPGA, the wall-clock time for PointNet and actor network inference are reduced by (45.26x, 60.66x), leading to 54.46x time savings.
Note that the performance models (Sec. \ref{sec:model-pointlk-core}--\ref{sec:model-reagent-core}) are able to predict the actual wall-clock time within 3\% error; their estimates are $C^\mathrm{L} / f = 23.84\text{ms}$ and $C^\mathrm{R} / f = 11.54\text{ms}$, whereas the actual runtimes are 23.23ms and 11.89ms for PointNetLK and ReAgent, respectively.

Fig. \ref{fig:ex5-m40-table} plots an example of the rotational ISO error over iterations.
We run the experiment on the proposed cores and embedded GPUs using a test sample from the ModelNet40 table dataset.
The full-precision PointNetLK converges to a reasonable solution after four iterations (96.02ms) on Xavier NX.
Compared to that, \textbf{PointLKCore} requires three more iterations to converge, which is possibly due to Jacobian matrices being affected by the quantization error, but takes only 14.54ms.
\textbf{ReAgentCore} takes four iterations (4.8ms) until convergence, which is 11.87x and 21.50x faster than Xavier NX and Nano.
While ReAgent runs faster than PointNetLK, it shows slight fluctuations after convergence.
ReAgent updates the solution by selecting a step size for each axis from a discrete action set, and the selected step does not always reduce the registration error as it may be slightly off from the optimal value.
PointNetLK treats the transform update $\vb{\Delta}\vb*{\xi}$ as a continuous variable, and the error monotonically decreases without noticeable oscillations.

\subsection{Jacobian Approximation Methods for PointNetLK} \label{sec:eval-pointlk-jacobi}
\begin{figure}[htbp]
  \begin{minipage}[b]{0.48\linewidth}
    \centering
    \includegraphics[keepaspectratio, height=2.1in]{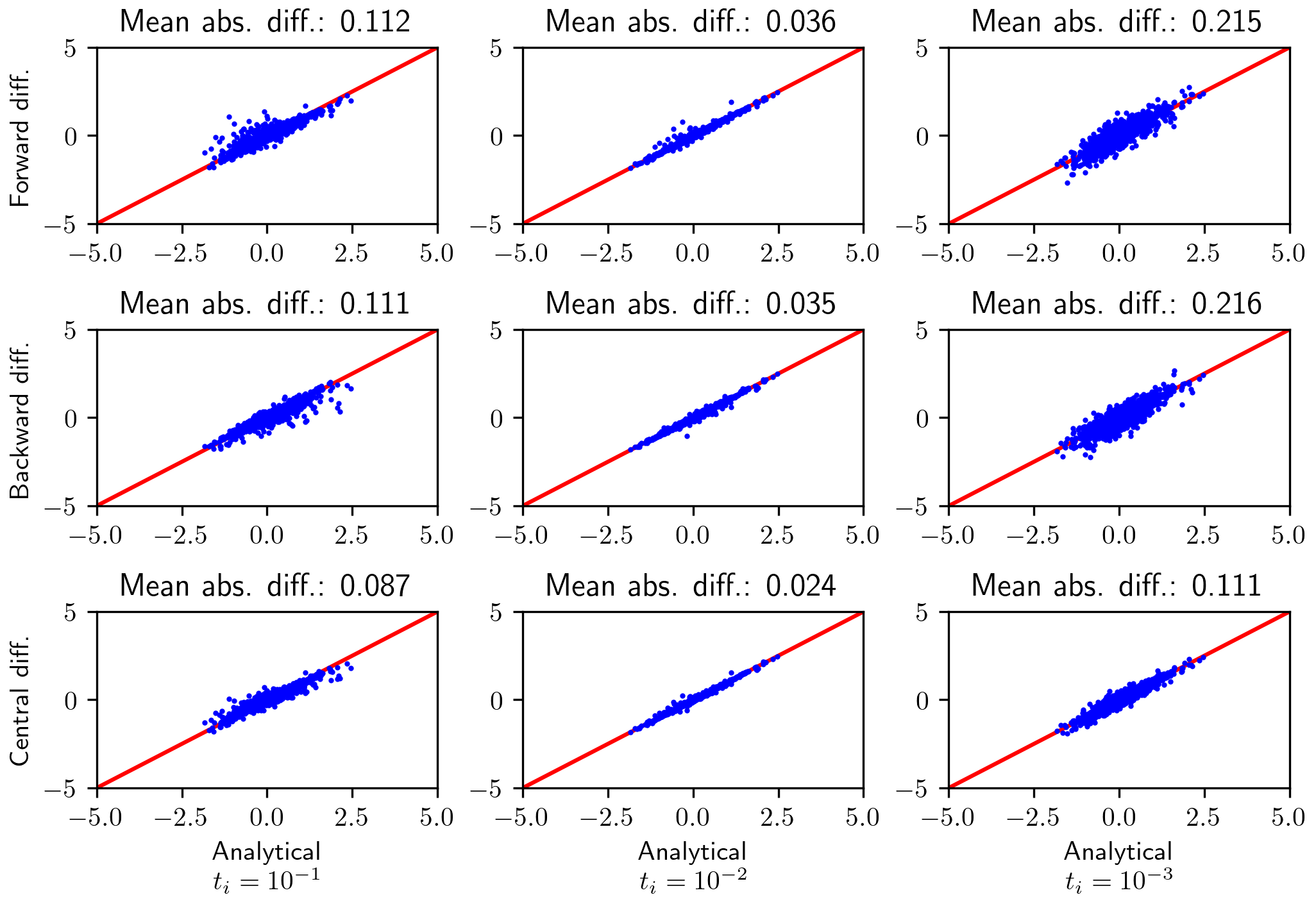}
    \subcaption{FP32}
    \label{fig:ex4-jacobi-fp32}
  \end{minipage}
  \begin{minipage}[b]{0.48\linewidth}
    \centering
    \includegraphics[keepaspectratio, height=2.1in]{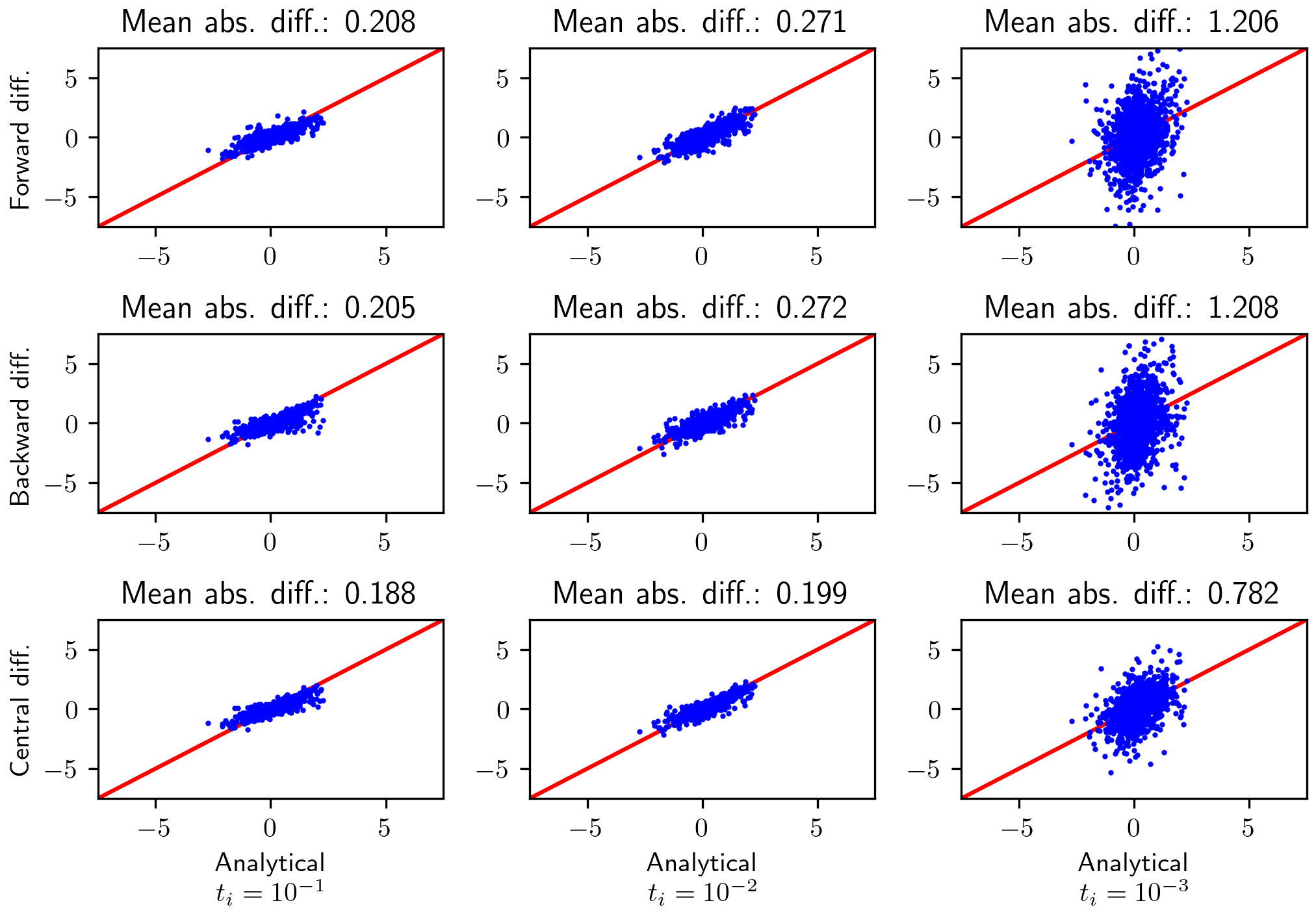}
    \subcaption{8-bit}
    \label{fig:ex4-jacobi-int8}
  \end{minipage}
  \caption{Difference between analytical and numerical Jacobians in PointNetLK.}
  \label{fig:ex4-jacobi-fp32-int8}
\end{figure}

Fig. \ref{fig:ex4-jacobi-fp32-int8} visualizes the numerical Jacobians (blue) in comparison with the analytical ones.
For simplicity, we focus on the third row of the Jacobian (i.e., the gradient of PointNet feature with respect to the rotation around $z$ axis) as in \cite{XueqianLi21}.
We run the full-precision and 8-bit PointNetLK with ModelNet40 (table category), and obtain numerical Jacobians using three finite difference approximations (forward, backward, and central).
The step size $t_i$ is varied from $10^{-3}$ to $10^{-1}$.
For the FP32 case (Fig. \ref{fig:ex4-jacobi-fp32}), the central difference with $t_i = 10^{-2}$ (bottom center) gives the best approximation with the minimum mean absolute error of 0.024 (blue dots are close to the red diagonal line).
While numerical Jacobians are affected by the 8-bit quantization (Fig. \ref{fig:ex4-jacobi-int8}), the central difference still improves the approximation quality and yields a comparable registration accuracy to the FP32 counterpart (Fig. \ref{fig:ex1-bits-and-jacobian}).
The numerical results show a noticeable deviation from the analytical ones when $t_i = 10^{-3}$, because quantization errors in the perturbed PointNet features (e.g., $\vb*{\phi}(\vb*{\delta} \vb{G}_i^+ \cdot \mathcal{P}_T)$) are amplified by the division by a small step $t_i$.
The analytical Jacobian incurs a significant increase in both computational and memory costs, in exchange for a slight improvement in the registration accuracy, which is due to the large feature gradient tensors (of size $(N, 3, 1024)$) involved during computation.
On ZCU104, PointNetLK (FP32) with the central difference takes 1.40s and has an ISO error of $0.939^\circ$, whereas that with the analytical Jacobian takes 22.4s (16.0x longer time) and gives almost the same error ($0.941^\circ$), indicating that the central difference is more suitable than the analytical solution.

\subsection{Power and FPGA Resource Consumption} \label{sec:eval-power-and-resource}
The power and energy efficiency of the registration is evaluated on the platforms listed in Table \ref{tbl:machines}.
While running a registration task with ModelNet40 (Unseen) and $N = 1024$, we collect the power consumption data at around 50ms interval and compute its average.
For a fair comparison, it is subtracted by the power consumption at idle state to exclude that of other peripherals such as LEDs and CPU fans.
We use \textit{tegrastats} utility for Nvidia Jetson boards.
On the desktop computer, we use \textit{s-tui} and \textit{nvidia-smi} for Intel CPU and Nvidia GeForce GPU.
For ZCU104, we utilize a Texas Instruments INA219 sensor and read out the current and voltage of the power supply.
The energy consumption per task is computed as a product of the average computation time (Sec. \ref{sec:eval-time}) and power consumption.
Table \ref{tbl:power-and-energy} shows the results.

On ZCU104, \textbf{PointLKCore} successfully reduces the energy per task by 36.96x with an additional power consumption of 0.68W, thanks to the 60.25x speedup.
As a result, it consumes (1.24x, 1.55x, 44.78x, 72.40x) less power and offers (10.90x, 25.39x, 114.50x, 163.11x) energy savings compared to running on the Xavier NX, Nano, desktop CPU, and desktop GPU.
In case of ReAgent, the proposed \textbf{ReAgentCore} reduces both power and energy per task by 1.47x and 80.15x, respectively, leading to (1.89x, 2.51x, 73.11x, 127.28x) and (20.96x, 52.44x, 173.12x, 231.58x) improvements in power and energy efficiency over these platforms.
FPGA-based ReAgent is 3.48x more energy-efficient than PointNetLK mainly due to 1.95x shorter runtime, while ReAgent has $0.33^\circ$ larger rotational error ($0.59^\circ$ vs. $0.92^\circ$).
Point-to-point ICP consumes less power and energy than \textbf{PointLKCore} when executed on Xavier NX, but its accuracy is considerably worse ($0.59^\circ$ vs. $4.89^\circ$) and is more affected by noise and initial rotations (Fig. \ref{fig:ex2-methods-m40}).
Compared to FGR, \textbf{PointLKCore} and \textbf{ReAgentCore} run with 3.01x and 10.46x less energy and are more robust to noise (Fig. \ref{fig:ex2-methods-m40}).
The results confirm the proposed cores yield up to two orders of magnitude savings in both power and energy costs, while maintaining the accuracy and robustness to noise.

\begin{table}[htbp]
  \centering
  \setlength\tabcolsep{3pt}
  \caption{Average power consumption and energy consumption per task}
  \label{tbl:power-and-energy}
  \begin{tabular}{l|rr|rrr|rrr} \hline
    & \multicolumn{1}{c}{ICP (pt2pt)} & \multicolumn{1}{c|}{FGR} &
      \multicolumn{3}{c|}{PointNetLK} & \multicolumn{3}{c}{ReAgent} \\
    & \multicolumn{1}{c}{CPU} & \multicolumn{1}{c|}{CPU} &
      \multicolumn{1}{c}{CPU} & \multicolumn{1}{c}{+GPU} &
      \multicolumn{1}{c|}{\textbf{+IP}} &
      \multicolumn{1}{c}{CPU} & \multicolumn{1}{c}{+GPU} &
      \multicolumn{1}{c}{\textbf{+IP}} \\ \hline
    \multirow{2}{*}{ZCU104} & 0.773W & 0.768W & 1.06W & -- & 1.74W & 1.43W & -- & 0.974W \\
      & 45.42mJ & 207.72mJ & 1489.92mJ & -- & \textbf{40.31mJ} &
      928.18mJ & -- & \textbf{11.58mJ} \\ \hline
    \multirow{2}{*}{Nano} & 1.53W & 1.64W & -- & 2.70W & -- & -- & 2.44W & -- \\
      & 45.81mJ & 213.58mJ & -- & 1023.56mJ & -- & -- & 607.30mJ & -- \\ \hline
    \multirow{2}{*}{Xavier NX} & 1.35W & 1.25W & -- & 2.16W & -- & -- & 1.84W & -- \\
      & 26.99mJ & 121.14mJ & -- & 439.54mJ & -- & -- & 242.74mJ & -- \\ \hline
    \multirow{2}{*}{Desktop} & 63.99W & 49.53W & 77.92W & 125.98W & -- & 71.21W & 123.97W & -- \\
      & 297.02mJ & 1372.55mJ & 4615.45mJ & 6575.14mJ & -- & 2004.74mJ & 2681.67mJ & -- \\ \hline
  \end{tabular}
\end{table}

\begin{figure}[htbp]
  \begin{minipage}{0.5\linewidth}
    \centering
    \setlength\tabcolsep{3pt}
    \captionof{table}{FPGA resource utilization (ZCU104)}
    \label{tbl:fpga-resource}
    \begin{tabular}[b]{c|rrrrr} \hline
      & BRAM & URAM & DSP & FF & LUT \\ \hline
      Total & 312 & 96 & 1728 & 460800 & 230400 \\ \hline
      \multirow{2}{*}{\textbf{PLKCore}} &
        174 & -- & 1341 & 147737 & 134298 \\
        & 55.77\% & -- & 77.60\% & 32.06\% & 58.29\% \\ \hline
      \multirow{2}{*}{\textbf{RACore}} &
        141 & 60 & 1264 & 133178 & 135155 \\
        & 45.19\% & 62.50\% & 73.15\% & 28.90\% & 58.66\% \\ \hline
    \end{tabular}
  \end{minipage}
  \begin{minipage}{0.48\linewidth}
    \begin{minipage}[b]{0.45\linewidth}
      \centering
      \includegraphics[keepaspectratio, height=1.4in]{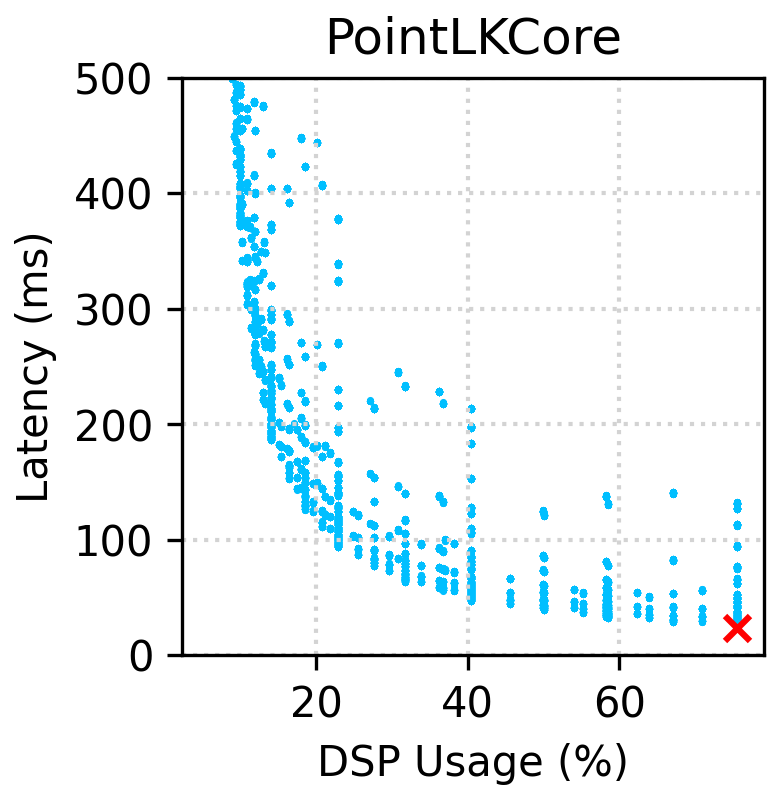}
    \end{minipage}
    \begin{minipage}[b]{0.45\linewidth}
      \centering
      \includegraphics[keepaspectratio, height=1.4in]{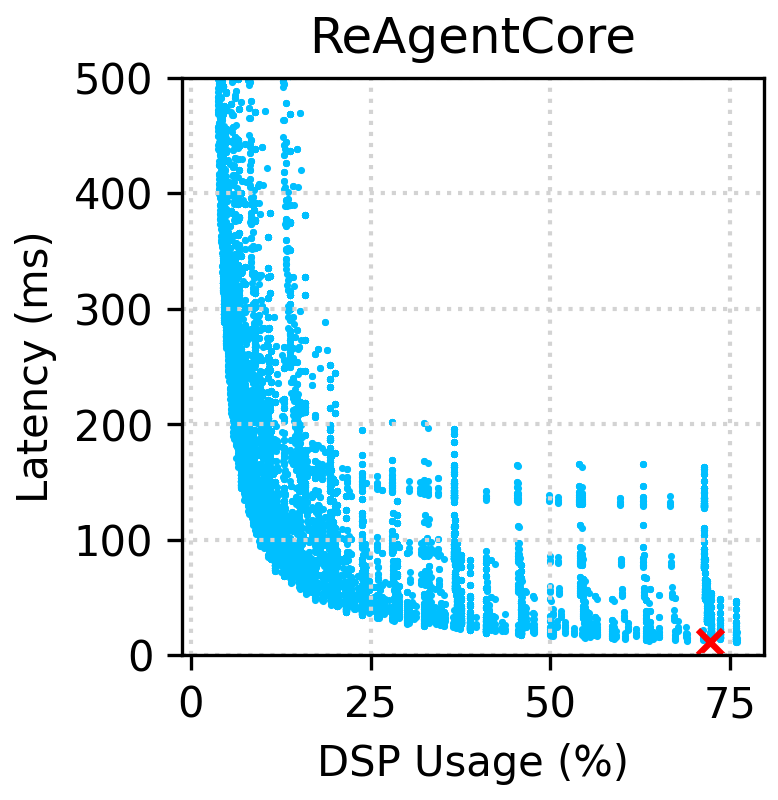}
    \end{minipage}
    \caption{Design-space exploration results.}
    \label{fig:dse-result}
  \end{minipage}
\end{figure}

Table \ref{tbl:fpga-resource} shows the FPGA resource utilization of \textbf{PointLKCore} and \textbf{ReAgentCore}, which are implemented with the design parameters in Table \ref{tbl:design-params}.
In the DSE process, the maximum resource utilization is set to 80\% to obtain synthesizable design points.
Both cores utilize more than 70\% of the DSP blocks to parallelize the computation in PointNet and actor networks.
Thanks to the 8-bit quantization and simple network architecture, the entire network parameters fit within the on-chip memory, and more than 40\% of the BRAMs are still available.
These BRAMs can be used to e.g., store input point clouds, which further reduces data transfer overhead from the external memory.
According to the resource models (Eqs. \ref{eq:model-pointlk-core}--\ref{eq:model-reagent-core}), the estimated DSP usage is $R_\mathrm{DSP}^\mathrm{L} = 1306$ (75.58\%) and $R_\mathrm{DSP}^\mathrm{R} = 1247$ (72.16\%) for \textbf{PointLKCore} and \textbf{ReAgentCore}, which are close to the actual results with an error below 2\%.
Compared to that, the BRAM usage is around 10\% less than the estimates ($R_\mathrm{BRAM}^\mathrm{L} = 217$ (69.55\%), $R_\mathrm{BRAM}^\mathrm{R} = 167.5$ (53.69\%)), as some on-chip buffers are implemented using FFs instead of BRAMs.
This overestimation does not negatively affect the performance, considering that our design is constrained by DSPs rather than BRAMs.
Fig. \ref{fig:dse-result} shows the correlation between the latency and DSP utilization ($C^\mathrm{L}$, $C^\mathrm{R}$, $R_\mathrm{DSP}^\mathrm{L}$, $R_\mathrm{DSP}^\mathrm{R}$ in Eqs. \ref{eq:model-pointlk-core}--\ref{eq:model-reagent-core}), obtained during DSE.
The red cross corresponds to the design point used for implementation (Table \ref{tbl:design-params}).
The result confirms that the promising design points are selected by DSE, under the objective of fully utilizing the FPGA resources and minimizing the latency.

Figs. \ref{fig:ex7-m40-half2-and-scnn} and \ref{fig:ex7-m40-half2-itr} show the qualitative results on the test samples taken from ModelNet40 (Unseen set) and ScanObjectNN.
While trained on synthetic point clouds, both \textbf{PointLKCore} and \textbf{ReAgentCore} successfully generalize to unseen object categories or real-world point clouds.
In addition, they find reasonable solutions within a few iterations and then refine the results in the subsequent iterations.

%% file: conc.tex

\section{Conclusion} \label{sec:conc}
This paper proposes a deep learning-based 3D point cloud registration for embedded FPGAs.
We design a fully-pipelined and parallelized PointNet feature extractor, based on which we develop two dedicated IP cores (PointLKCore and ReAgentCore) for the recently-proposed iterative methods: PointNetLK and ReAgent.
By simplifying the PointNet architecture and processing input point clouds in small chunks, the on-chip memory cost becomes independent of input size, leading to the resource-efficient design.
We apply the hardware-friendly LLT quantization for PointNet and actor networks, which only involves table lookup operations during inference.
The whole network fits within on-chip memory as a result and the data transfer overhead is minimized.
To further improve the accuracy of PointNetLK, we propose to use the central difference approximation for Jacobians and train the model jointly with a decoder or classifier.
We conduct the design space exploration based on the latency and resource models to fully exploit the computing power of FPGAs.

The proposed cores provide favorable accuracy and speedup on a wide range of input size, compared to their software counterparts and classical approaches.
They are more robust to large initial misalignments and noise than ICP and FGR as they do not rely on correspondences or hand-crafted features, and generalize well to unseen object categories or real-world point clouds.
The experimental results highlight the characteristics of two methods as well; PointNetLK is more stable and accurate in case of small initial rotations, while ReAgent converges in fewer iterations.
On ZCU104, PointLKCore and ReAgentCore find reasonable solutions in less than 15ms and run 45.75x and 44.08x faster than ARM Cortex-A53 CPU.
They achieve 2.64--7.83x and 1.98--11.13x speedup over Intel Xeon CPU and Nvidia Jetson devices, consume less than 1W, and are 163.11x and 213.58x more energy-efficient than Nvidia GeForce GPU.
These results indicate that the FPGA-based custom accelerator is a promising approach compared to using embedded GPUs or desktop CPUs to tackle the computational complexity of learning-based registration.
In future work, we aim to extend this work to address more complex tasks such as object tracking and SLAM.
Other network architectures could be employed instead of PointNet to extract more distinctive features and further improve the accuracy.